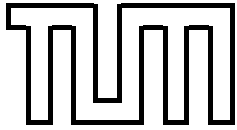

TECHNISCHE UNIVERSITÄT MÜNCHEN

FAKULTÄT FÜR INFORMATIK

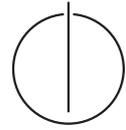

**Lehrstuhl für Echtzeitsysteme und Robotik**

# Implementation and Evaluation of multimodal input/output channels for task-based industrial robot programming

## Implementierung und Evaluierung multimodaler Ein-/Ausgabekanäle zur aufgabenbasierten Programmierung industrieller Roboter

Masterarbeit in Robotics, Cognition, Intelligence

Author:              Stefan Profanter
Supervisor:          Prof. Dr.-Ing. Alois Knoll
Advisor:             Alexander Perzylo
Submission Date:     15.10.2014

Ich versichere, dass ich diese Masterarbeit selbständig verfasst und nur die angegebenen Quellen und Hilfsmittel verwendet habe.

*I assure the single handed composition of this master thesis only supported by declared resources.*

München, den 15.10.2014

Stefan Profanter

# Abstract

Industrial robots are used in many different fields of application: from huge welding production lines for automobiles to assembly tasks in Small and Medium-sized Enterprises (SMEs). Such robot systems require specially trained employees, who know how to program these robots and how to teach them new tasks using TeachPads and specific programming languages like KRL (Kuka) or RAPID (ABB).

Programming these robots is not very intuitive, and the programmer has to be a domain expert for e.g. welding and programming to know how the task is optimally executed. For SMEs such employees are not affordable, nor cost-effective. Therefore a new system is needed where domain experts from a specific area, like welding or assembly, can easily program a robot without knowing anything about programming languages or how to use TeachPads. Such a system needs to be flexible to adapt to new tasks and functions. These requirements can be met by using a task based programming approach where the robot program is built up using a hierarchical structure of process, tasks and skills. It also needs to be intuitive so that domain experts don't need much training time on handling the system. Intuitive interaction is achieved by using different input and output modalities like gesture input, speech input, or touch input which are suitable for the current task.

This master thesis focuses on the implementation of a user interface (GUI) for task based industrial robot programming and evaluates different input modalities (gesture, speech, touch, pen input) for the interaction with the system. The evaluation is based on a user study conducted with 30 participants as a Wizard-Of-Oz experiment, where non expert users had to program assembly and welding tasks to an industrial robot, using the previously developed GUI and various input and output modalities.

The findings of the task analysis and user study are then used for creating a semantic description which will be used in the cognitive robotics-worker cell for automatically inferring required system components, and to provide the best suited input modality.

# Zusammenfassung

Industrieroboter haben ein weitreichendes Anwendungsgebiet: von großen Schweiß-fließbändern für die Fahrzeugindustrie bis hin zu Montageaufgaben in Kleinen und Mittelständischen Unternehmen (KMU). Diese Robotersysteme erfordern speziell ausgebildetes Personal für die Programmierung der Roboter unter Verwendung von TeachPads und speziellen Programmiersprachen wie KRL (Kuka) oder RAPID (ABB).

Die Programmierung dieser Roboter ist weder intuitiv noch schnell erlernbar; zusätzlich sollte der Programmierer ein Fachexperte für z.B. Schweißen sein, um eine effiziente und korrekte Ausführung der Aufgaben zu gewährleisten. Für KMUs sind Angestellte für nur diesen Zweck weder finanziell tragbar noch kosteneffizient. Aus diesem Grund werden neue Systeme benötigt, bei dem Fachexperten für z.B. Schweißen oder Montage ohne spezielles Vorwissen im Bereich der Roboterprogrammierung in der Lage sind, diese Industrieroboter zu programmieren. Solche Systeme haben eine hohe Anforderung an die Flexibilität und Anpassung an neue Aufgaben. Diese Anforderungen können durch Verwendung von aufgabenbasierter Programmierung erreicht werden, bei der das Programm hierarchisch in Prozesse, Aufgaben und Fähigkeiten aufgeteilt ist. Das System muss zusätzlich intuitiv bedienbar sein, um aufwändige Schulungen zu vermeiden. Intuitive Interaktion wird erreicht durch verschiedene Ein- und Ausgabekanäle wie z.B. Gesten, Spracheingabe oder Touch Eingabe passend für die aktuelle Aufgabe.

Der Schwerpunkt dieser Masterarbeit liegt in der Implementierung einer grafischen Benutzeroberfläche für die aufgabenbasierte Programmierung von Industrierobotern und der Evaluierung verschiedener Eingabekanäle (Gesten, Sprache, Touch, Stifteingabe) zur Interaktion mit dem System. Die Evaluierung basiert auf einer Benutzerstudie mit 30 Teilnehmern, durchgeführt als ein Wizard-of-Oz Experiment. Dabei mussten die unerfahrenen Benutzer verschiedene Aufgaben wie z.B. Schweißen oder Montage unter Verwendung der grafischen Benutzeroberfläche und verschiedenen Eingabemodalitäten dem Roboter beibringen.

Die Ergebnisse der Aufgabenanalyse und der Benutzerstudie werden anschließend für die Erstellung einer semantischen Beschreibung verwendet, die in der kognitiven Roboter-Arbeiter Zelle zur automatischen Inferenz der erforderlichen Systemkomponenten und zur Vorauswahl der am besten geeigneten Eingabemodalität verwendet werden kann.

# Contents

# 1 Introduction

Since the first industrial robots were developed in the 1970s, the research in intuitive robot programming got more an more important. Large companies with a lot size of a few thousand can afford specially trained robot programmers for their manufacturing systems, but for Small and Medium-sized Enterprises (SMEs), buying a robot is still something which isn't profitable. Current industrial robot systems are complex to program. SMEs normally have small lot sizes and don't want to have big stocks. Consequently, concepts like flexibility and agility are fundamental in actual manufacturing plants [JN Pires, 2006]. [Wallhoff et al., 2010] states that "Flexibility and adaption to rapidly changing market demands is one of the highest design criteria for the working cells and production facilities of the future."

Therefore research groups all over the world try to build systems which combine hybrid assembly (collaboration between human and robot) and easier programming, even with no special knowledge in industrial robot programming. By using different input modalities, not specially trained employees should be able to work with the robot and reprogram/adapt the robot to a new production cycle. Relatively few recent robotic systems are currently equipped with multimodal user interfaces that permit to program the robot, or simply control the robot using natural means [Burger, Ferrané, and Lerasle, 2010]. A multimodal system is "a system that features multiple input devices (multi-sensor interaction) or multiple interpretations of input issued through a single device" [Chatty, 1994].

The goal of this thesis is to develop a multimodal user interface (UI) for industrial robot programming, used for industrial manufacturing tasks, and evaluate the multimodal interaction with the system. The focus hereby lies on the evaluation of multimodality using different input methods such as touch, gestures and augmented reality. The findings from task analysis and the evaluation are then used to create a semantic description which can be used within cognitive human-robot worker cells.

## 1.1 Motivation

Current industrial robots require a fairly complex and time consuming teach-in process for each task it has to execute. New tasks are defined using TeachPads, or by manually specifying each coordinate and each single step of the task. Using such a TeachPad requires special training and a lot of experience and therefore they are not very intuitive. Using a multimodal approach to program these robots would make the teach-in process less complicated. A multimodal interface can include gestures, speech, haptics, eye blinks, touch input and many others. Such multimodal interfaces have many advantages: they are more robust against errors and even prevent many errors due to the natural interaction with the system. They help users to correct errors or recover from them more easily and bring more bandwidth to the communication. Multimodal systems can adapt easier to different situations and environments. [Jaimes and Sebe, 2007] Among those advantages, flexibility offers the user to select the most adequate modalities in his specific situation, according to his preferences and capabilities. Multimodal interfaces permit diverse user groups to control on how they interact with the system and therefore accommodate a broader range of users than traditional interfaces - including users of different ages, skill levels, native language or even temporary illnesses or permanent handicaps. Multimodal interfaces even improve the efficiency by more than 10% to 25% when using multiple modalities at the same time. [Oviatt, 2012]

## 1.2 Industrial robot characteristics

A robot in general is a system or machine guided by a controller or electrical circuit having actors to interact with the world. It is in most of the cases equipped with sensors to perceive the surrounding environment and act based on the current world state. Industrial robots are a specific subclass of robots located at a fixed position, repeating tasks over and over again while providing the possibility to reprogram and perform new tasks. The main difference to other numerically controlled machines is the versatility of the robot: it can be equipped with different tools and has a much larger workspace compared to the volume of the robot itself.

The Robot Institute of America uses the following definition:

> A robot is a re-programmable multifunctional manipulator designed to move materials, parts, tools or specialized devices through variable programmed motions for the performance of a variety of tasks. [Tzafestas, 2013]

The most important statement in this definition is the word "re-programmable" which gives the robot two main characteristics: namely utility and adaptability. Those two characteristics are very important for industrial robots, and even more when used in SMEs (Small and Medium-sized Enterprises) [Wallén, 2008].

The International Organization for Standardization, ISO, uses another definition for a robot, "A manipulating industrial robot is an automatically controlled, re-programmable, multi-purpose, manipulative machine with several degrees of freedom, which may be either fixed in place or mobile for use in industrial automation applications" [Tzafestas, 2013].

In summary, an industrial robot shall be easily re-programmable without modifying its physical structure. The robot should also be able to work independently and autonomously, and it should be flexible to provide easy adaption to new tasks.

## 1.3 History of industrial robots

The first industrial robot as we know it today was developed in 1954 by George Devol, who patented the first multijoined robotic arm called *Programmed Transfer Article*, but he was uncertain on how the machine can be used. In 1956 Devol met Joseph Engelberger and founded the new company Unimation (from Universal Automation). They visited 15 different car factories and around 20 other industries to determine the needs regarding industrial robots. The result of this gathering of requirements was the robot Unimate which was released in the year 1961 and installed in a General Motor's (GM) factory to serve a die casting machine (see Figure 1.1). It was a fairly simple robot able to perform only one task [Westerlund, 2000].

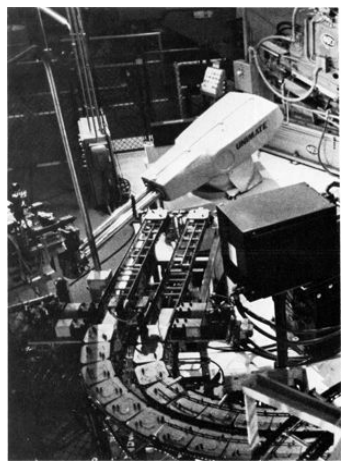

**Figure 1.1:** *The first industrial robot, Unimate, sequenced and stacked hot pieces of die-cast metal. General Motors (GM) factory in 1961. Source: http://www.prsrobots.com/1961.html*

After installing another 66 robots for GM, Devol and Engelberger were convinced that the industrial robot had a future, but at this time the market wasn't ready yet: the manufacturing industry wasn't very interested in industrial robots, whereas media was the opposite. Engelberger, Devol and their robots were regular guests on the American television, where the robots served coffee or appeared in commercials where they served beer. The robot changed to be a "funny toy" instead of a scary object [Wallén, 2008].

The first robots in Europe were installed in the year 1967 at Svenska Metallverken in Upplands, Sweden. Those robots did simple monotonous tasks like picking up objects and placing them somewhere. These simple pick and place robots were followed by robots for spot-welding, which were installed in 1969 at GM for spot-welding car bodies, and in 1972 at Fiat, where the first spot-welding line was set up [Westerlund, 2000].

Victor Scheinmann, a mechanical engineering student working in the Stanford Artificial Intelligence Laboratory (SAIL), designed in 1969 the Stanford Arm. This 6-DOF (degrees of freedom) all-electric robot manipulator introduced the non-anthropomorphic kinematic configuration, which made solving the equations for robot kinematics easier and therefore increased the computation speed. Subsequent robot designs were strongly influenced by Scheinman's concepts [Hägele, Nilsson, and Pires, 2008].

Trallfa, a norwegian company, saw the potential of the robots and wanted to install a Unimate robot for painting their wheelbarrows as a flexible spray painting device due to the bad working environment and thereby difficulties to recruit new personnel. The Unimate robot was too expensive, thus they developed a robot on their own costing only 15 000 Norwegian crowns (the Unimate would have cost 600 000 Norwegian crowns). In 1967 they presented their electro-hydraulic easy to program robot. In 1985 ASEA (later ABB) took over Trallfa and integrated the painting robots into their industrial robot portfolio [Westerlund, 2000].

In 1973 there were already 3000 robots in operation around the world, 30% of them were Unimation robots [Hägele et al., 2008; Wallén, 2008].

The first microcomputer, the Intel 8008, was built into ASEA's IRB 6, their first prototype of an industrial robot. Due to the limited space of 8KB the program was extremely complicated. The IRB introduced some new concepts such as fully electrical (drive and control systems), an anthropomorphic structure and the use of microcomputer. The first customer of IRB 6 was a small company with only 20 employees: Magnussons i Genarp AB. They used the robot for stainless steel pipe production for the food industry. Using these robots he was the first in the world to operate an unmanned factory 24/7. The IRB 6 has proved to be very robust: life-times of more than 20 years in harsh productions were reported.

The PUMA (programmable universal machine for assembly, 1979) from Unimation became one of the most popular robot arms due to its dexterity which was close to the one of a human arm [Hägele et al., 2008; Westerlund, 2000].

In the end of the 1970s and the beginning of the 1980s, the development of robots was mainly concentrated on assembly tasks. Arc welding on the other hand required better motors and control systems. Takeo Kanade, a japanese computer scientist, solved this problem by building a direct drive arm, where motors are directly installed into the joints of the arm. In the mid 1980s, industrial robots were used for machine tending, material transfer, painting and welding.

In 1978, the *selective compliance assembly robot arm* (SCARA) was developed. It is particularly suited for assembly tasks due to the rigidity in the vertical axis and compliance in the horizontal axis. Its low cost design and fast motions contributed significantly to create a world-wide boom in high-volume electronics production and consumer products (see Figure 1.2).

In 1986, Honda began a robot research program that started with the premise that the robot "should coexist and cooperate with human beings, by doing what a person cannot do, and by cultivating a new dimension in mobility to ultimately benefit society."[Donohoe, 2007]

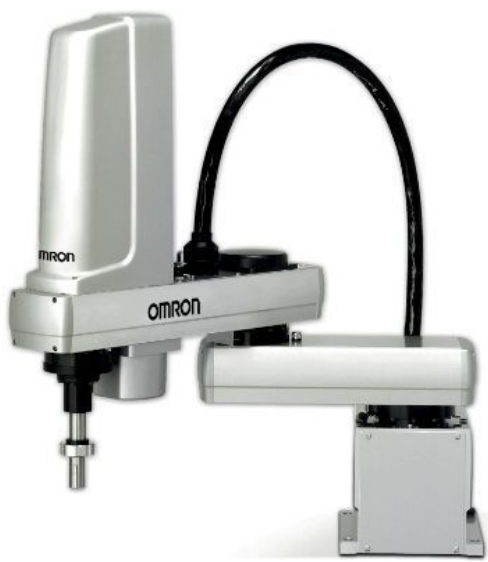
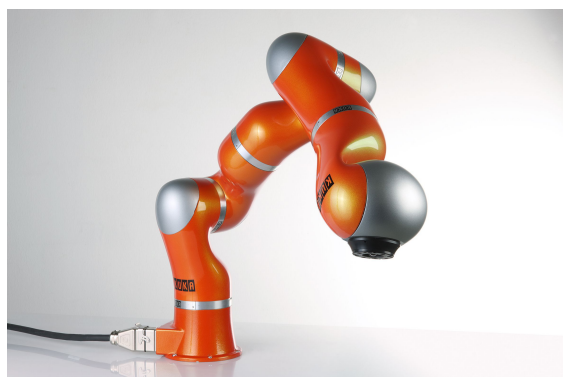

(a) SCARA                    (b) Kuka lightweight robot

**Figure 1.2:** *(a) The SCARA (selective compliance assembly robot arm) has a high rigidity in the vertical axis and compliance in the horizontal axis. Source: http://industrial.omron.de. (b) The Kuka lightweight robot has a weight-to-load ratio of 1:1. Source: http://www.metal-supply.com.*

From early on, the reduction of the mass and inertia was a primary target in robot research with the ultimate goal of a weight-to-load ratio of 1:1, similar to the human arm. KUKA reached this goal in the year 2006 by presenting the 7-DOF KUKA lightweight robot (see Figure 1.2).

Apart from the traditional cartesian design, the company Güdel introduced a curved-track gantry in 1998. This new design allowed the robot to circulate and change its location.

In 2005 the first commercial robot for synchronized, two-handed manipulation was introduced by MOTOMAN. This system features 13 axes of motion and is ideal for placing it on a site that was previously accommodated by human workers.

In the following years research mainly focused on simultaneous and synchronous operation of robots on a single workpiece and the increased usage of vision systems for object identification, localization, and quality control.

Decreasing cost of robot systems and advancement in automation flexibility makes the usage of industrial robots for Small and Medium-sized Enterprises (SMEs) more and more attractive. Unfortunately these systems are not yet flexible and intuitive enough to achieve a widespread use for SMEs.

## 1.4  Programming concepts for industrial robots

The development of modern motion-oriented robot programming languages dates back to the mid 1970s. One of the first languages was VAL (Variable Assembler Language), used for Unimation industrial robots. It uses a clear and self explanatory instruction set. VAL also supports libraries and routines which allow the programmer to create subtasks and reuse code. An example VAL program is shown in Listing 1.1. Due to it's simplicity it was fairly impossible to specify complex arithmetic computations or to use complex sensor data. VAL was also used for programming PUMA robots. Stäubli currently uses VAL3 for their robots.

```
PROGRAM PICKPLACE
   1.  MOVE P1
   2.  MOVE P2
   3.  MOVE P3
   4.  CLOSEI 0.00
   5.  MOVE P4
   6.  MOVE P5
   7.  OPENI 0.00
   8.  MOVE P1
.END
```

**Listing 1.1:** *A simple VAL program moving the robot to a save position (P1), then to approach position (P2), moving to the object (P3), closing the gripper and placing the object on position (P5). (Source: http://en.wikipedia.org/wiki/Robot_software#Examples_of_programming_languages_for_Industrial_Robots)*

Already in the early 1970s reasearch groups started to focus on so called task-oriented programming languages. One of the first examples are IBM's AUTOPASS system and the LAMA system proposed by MIT. The basic idea behind this approach is to make robot programming easier by relieving the programmer from knowing all the specific hardware details, achieved by avoiding to code every tiny motion/action. Instead the programmer is specifying in an intuitive way what the robot should do and how it should do it.

It turned out that the most difficult part in automatic and task based robot programming is the *execution and control* of automatically generated action/motion sequences and automated generation of collision-free paths. Besides some simple applications, like 4-DOF assembly of PCBs, this is still a subject of current research. [Wahl and Thomas, 2002].

Current programming concepts can be divided into two main categories: *Online programming* and *Offline programming*:

### 1.4.1 Online programming

In online programming, the program is written directly on the robot system or by using the robot itself. The most popular method is Teach-In where the programmer can use a so called TeachPad (see Figure 1.3) which allows direct robot controlling by moving the robot through a 6D mouse and touch input. With the TeachPad the programmer can generate new programs by inserting the instructions and defining the position and pose for each command moving it to the specific position. These programs are still text based and require sufficient knowledge about the used programming language.

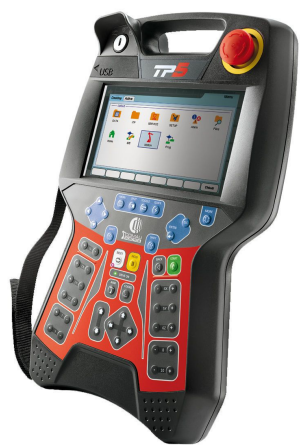

**Figure 1.3:** *A TeachPad used for Teach-In online robot programming. Source: http://www. directindustry.com*

### 1.4.2 Offline programming

Offline programming has the main advantage that you don't need the robot to write programs. The programmer writes the program on a PC which is independent of the robot system. When finished the program is uploaded to the robot and executed directly.

The most commonly used offline programming concept is text-based programming: tasks are defined on the basis of problem oriented modern programming languages. The advantage compared to online programming is that the program can be easily changed and documented. Unfortunately each robot manufacturer has its own programming language which requires a specially qualified programmer.

CAD based program specification is another offline programming concept where robots are programmed based on technical drawings an simulations. The programmer can use a 3D software in which the robot and workpiece are shown as 3D models and the robot can be moved and positions can be programmed. This allows to detect planning and construction errors early and the scene can be viewed from any position. The disadvantage is that all the models of the workpiece, robot and tools must be identical to the real one to avoid errors. Flexible pipes are also hard to visualize within the simulation but are an important factor when planning robot movements because they result in significant movement restrictions. Additionally you need to calibrate the robot to fit the simulated world to the real world.

## 1.5 Contribution

This master thesis contributes to the field of intuitive robot programming by evaluating different input modalities for task-based robot programming and by providing an ontology which can be used by cognitive robot cells for semantic reasoning about input modalities and task parameters. The evaluation is performed through a user study conducted with 30 participants using a newly developed touch-friendly user interface which integrates the four input modalities: speech, gesture, touch and pen.

Chapter 2 (Related Work) gives an overview over previously published work for intuitive robot programming, human-robot interaction, multimodal systems and user studies for multimodal interaction. In Chapter 3 (Input & Output Modalities) different human input and output channels are described which can be used for interacting with computer systems through computer input and output modalities. It also includes a historical overview of the different modalities and evaluates advantages and disadvantages. Chapter 4 (Domains & Tasks) introduces the hierarchical structure for task-based robot programming using Process, Task and Skill notation. This chapter describes some basic skills required from a robot system and lists example tasks including their parameters for different domains like assembly, welding, woodworking or metal processing. The input modalities, Skills, Tasks, Parameters and system components are then modeled within an ontology. This ontology maps abstract descriptions of tasks to basic instructions for the robot and describes causal relations between low level system components like sensors and higher level task descriptions.

Chapter 5 (Graphical User Interface) gives an overview over the implementation of the graphical user interface developed for task-based robot programming with the main goal to provide an easy understandable and intuitive interface for domain experts to program a robot. The used technologies and frameworks are listed and the basic structure of the software is described, which was built up to allow easy integration of new software components and quick adaption to new functions and requirements. This chapter also includes a description how this GUI was integrated into the already existing system using ROS (Robot Operating System).

Chapter 6 (User Study) describes the structure of the user study which was conducted to evaluate different input modalities for task based robot programming. It describes the four main phases of the study and how the study was conducted using the Wizard-of-Oz approach. In Chapter 7 (Evaluation) follows the evaluation of the user study and analysis of the results. The results of the study are analyzed and assessed.

The last Chapter 8 (Conclusions) summarizes the results of this master thesis and highlights the outcomes of the user study. It also gives an outlook for further improvements of the system and challenges for further research projects.

# 2 Related Work

This master thesis is related to different fields of research. Multimodal input is used for human-robot interaction to provide an intuitive interface for task-based programming and interacting with robots. A user study is then conducted, which evaluates previously defined input modalities and the multimodal system in general.

The following sections give a brief overview on related work in those areas by summarizing previous research projects and publications.

## 2.1 Multimodal interaction

A review of major approaches in multimodal human-computer interaction, especially for body, gesture, gaze and affective interaction is given in [Jaimes and Sebe, 2007]. They state that most researchers process different channels (visual, audio) independently. Additionally multimodal fusion is still in its infancy and a combination of low-level features, high-level reasoning, and natural language processing is likely to provide the best emotion inference in the context of MMHCI (multimodal human-computer interaction). A more comprehensive survey of existing and recent advances in the field of HCI can be found in [Karray, Alemzadeh, Saleh, and Arab, 2008].

[Ruiz, Chen, and Oviatt, 2010] give an overview over the advantages of multimodal input interfaces (Robustness, Naturalness, Flexibility, Minimising errors) and evaluate cognitive load and performance in relation to multimodal interfaces. [Hinckley and Wigdor, 2012] and [Oviatt, 2012] provide a comprehensive overview on different input modalities and evaluate these based on various aspects like input speed or user feedback possibilities and clear up some myths about multimodal interfaces by discussing how users normally react to them.

[Kawamura, Nilas, Muguruma, and Adams, 2003] presents an agent-based architecture for mixed-initiative interaction between a human and a robot. The platform provides a basis for developing various agents that control robots and user interface components. This agent-based approach allows an efficient interaction between human and robots through an adaptive interface where an *Adaptive User Interface Manager* shows only the most relevant information from each agent for the current mission.

In [Iba and Paredis, 2004] a multimodal robot programming framework is described that allows the user to program a robot interactively through an intuitive interface using hand gesture and spontaneous speech recognition. [Hahn, 2010] examines different new input modalities with a focus on gesture and speech input and concludes that there's still some work to do especially in the sector of human voice detection.

## 2.2 Intuitive programming and interaction with robots

When working together with a robot, multimodal interaction can be used to instruct the robot to perform specific tasks or support the human worker. [Bannat, Gast, Rehrl, and Rösel, 2009] present a system where the human can give commands via three different input modalities (speech, gaze and soft-buttons). They show that it is possible to use multimodality in a working hybrid assembly process where humans are working together with an industrial robot.

[Wahl and Thomas, 2002] describe a system which allows intuitive task-oriented robot programming using symbolic spatial relations (SSR), defined in [Ambler and Popplestone, 1975]. SSRs describe relations between assembly parts, e.g. "shaft1 of bulb fits hole1 of rack". They state that there's still a tremendous gap between commercially available robot programming languages and methods developed in research laboratories all over the world.

A further step into more intuitive teaching methods for small and medium enterprises is described in [Schraft and Meyer, 2006]: they present a method for adapting trajectories by transforming them into sequences of geometric primitives. Then these primitives can get changed according to the specific task resulting in a significantly faster teach-in process. Another approach of defining welding and gluing trajectories is described in [Meyer, Hollmann, Parlitz, and Hägele, 2007] where the user can program the robot by demonstration supported by graphical and verbal interfaces. [J.N. Pires, Godinho, and Araújo, 2007] explores the possibility to obtain robot programs from technical drawings on a sheet of paper using a digital pen in combination with CAD models. He shows that this approach is very useful and powerful, especially for Small and Medium-sized Enterprises.

Another CAD-based human-robot interface allowing non-expert users to teach a robot, is presented in [Neto, Mendes, Araújo, Pires, and Moreira, 2012]: they show that it is possible to generate robot programs from common CAD drawings and run them on the robot. The programmer opens a 3D CAD Drawing and can draw robot trajectories directly into the model.

In [Carø e, Hvilshø j, and Schou, 2014] an intuitive programming system for a mobile

KUKA LWR is developed which uses a hierarchical structure (task, skill and device primitives) to generate a robot program. This tool is called *Little Helper ++*. The developed Terminal and Graphical User Interface still requires the user to define a large number of parameters since the system doesn't use any knowledge base for automatic reasoning. They conclude that the skill-based approach is beneficial for programming and executing complex industrial tasks, especially when trying to abstract from complex low level robot functionalities as used for task-based robot programming.

[Neto, Pires, and Moreira, 2010] describe a system using a hand-held accelerometer (Wii Remote Controller) in combination with speech input for high-level robot programming. They use Artificial Neural Networks (ANNs) for gesture recognition to control the movement of the robot and reach a recognition rate up to 96%.
Combining different input modalities with augmented reality (AR) even improves the intuitiveness of human-robot interaction. [Ameri, Akan, and Cürüklü, 2010] use images from a camera mounted on the gripper enhanced with information overlays and shows that it helps human workers to operate and program a robot easier and faster.
[Gaschler, Springer, Rickert, and Knoll, 2014] present in their paper a novel augmented reality system for defining virtual obstacles, specifying tool positions, and specifying robot tasks. A 3D input device is used to define points in real world, an overlay projection and monitor give the user feedback on the current system state.

A hybrid assembly station for human-machine coworker is described in [Wallhoff et al., 2010]: The robot learns new tasks from worker instructions using voice, gaze and tactile interaction backed by a knowledge-based system controller that is orchestrating the skills of connected modules and is representing sensor events and the work tasks as first order logic predicates. [Rigoll, 2011] states that such hybrid assembly stations require complete repertoire of multimodal recognition techniques, advanced output generation, as well as intelligent knowledge processing in order to be able to assist human workers in assembly tasks with and without additional support from robots.
For robots who collaborate with human workers it's important to know which action the human is currently performing and if it is safe for the robot to move. [Hartmann, 2011] developed for his PhD thesis a system which uses visual trackers and inertial measurement units and is able to recognize low-level activities and complex tasks which are sequences of low-level actions. He demonstrated that for task level recognition the incorporation of context knowledge is essential for recognition from uncertain inputs.
Another gesture based control and programming system is proposed by [Lambrecht, Kleinsorge, and Kruger, 2011] where the user can define trajectories using hand gestures and program an assembly sequence through task demonstration [Lambrecht, Kleinsorge, Rosenstrauch, and Kruger, 2013]. Lambrecht constructs a virtual scene reconstruction and pose estimation of the objects using 2D cameras from which the subsequent poses of the objects can be derived. Out of these subsequent poses, a robot program is generated.
[Gleeson, MacLean, Haddadi, Croft, and Alcazar, 2013] produced a gestural lexicon by observing human interactions during an industrial based assembly task. The gestures were then implemented on a robot arm and evaluated in human-robot trials. Glesson concludes that the gestures they found are intuitive to understand, even if they are executed by a robot arm.

Most of the systems presented above are optimized for a specific domain or task and are not very flexible or adaptive. Using those systems often requires special knowledge about robotics and is therefore not suitable for domain experts of other areas working in Small and Medium-sized Enterprises. By integrating knowledge databases, most of these parameters could be inferred automatically making it easier to program the system and to be more flexible and adaptive. For some of the projects the graphical user interface is not that intuitive or the system heavily relies on speech input which isn't appropriate for usage in loud environments like assembly factories.

## 2.3  User study

Designing a user study is a research field on its own. There are various ways a questionnaire can be structured or how to focus on different aspects. A comparison of different survey types and implementations in human-computer interaction is given in [Ozok, 2012]. Ozok lists and compares various survey types, design techniques and evaluation techniques. He also discusses implementation challenges for user studies in open and controlled environments.
[Kühnel, Westermann, Weiss, and Möller, 2010] compare in their publication established questionnaires and interaction parameters and evaluate different rating methods for usability measurements (SUS, CSUQ, SUMI) and to measure intuitive use and user interface satisfaction (QUIS). They state that AttrakDiff, SUS and USE are suitable for usability evaluation of systems with multimodal interfaces but the selection of an appropriate questionnaire depends on the goal of the evaluation.

[Dahlbäck, Jönsson, and Ahrenberg, 1993] argument in their paper "Wizard of Oz Studies - Why and How" that Wizard of Oz studies are required to consider the unique qualities of man-machine interaction. Dialogs between human and human differ from dialogs between human and machine and therefore it may have a major influence if a user talks to a machine instead of a human which makes data from human-human interaction an unreliable source for designing natural language processing systems. Dahlbäck describes different factors which influence how we speak to a human or a machine an so it's necessary to use Wizard of Oz studies to fully evaluate the requirements for the system.

SUXES is a user experience evaluation method for spoken and multimodal interaction described in [Turunen and Hakulinen, 2009]. It captures user expectations and user experiences, making it possible to analyze the state of the application and its interaction methods and compare the results. It's suitable for iterative development and prototyping, while the most important feature is the collection of user expectations.

## 2.4 Evaluation of multimodal systems

There already are different user studies and evaluations of multimodal systems for various types of tasks or domains. This section presents some of these studies and highlights their result related to multimodal input and output.

[Ren, Zhang, and Dai, 2000] found out, that for CAD applications, users prefer pen+speech+mouse input. Within another independent study by Ren, users preferred pen+speech for map applications. Not only the type of the task but also the complexity influences the choice of modality: [Oviatt, Coulston, and Lunsford, 2004] reports that the likelihood of multimodality increases with the cognitive load of the task.

[Bellik, Rebaï, and Machrouh, 2009] show that the output modality used by the system influences the choice of the user too: if the system uses verbal output, users tend to use speech input, while graphics output (icons) lead to touch screen use. Another influencing factor is the situational context in terms of privacy of the situation, where the interaction takes place. [Jöst, Häuß ler, Merdes, and Malaka, 2005] and [Wasinger and Krüger, 2006] proof a declining willingness to use multimodal input with decreasing intimacy of the relationship between the user and other persons around and that they prefer to use non-observable modalities like handwriting and touch gestures.

The computer literacy (expert vs. non expert users) additionally influences the choice of modality. Expert users prefer to interact multimodally, whereas non expert users showed no preference between unimodal and multimodal [Angeli and Gerbino, 1998]. Similar results were gained by [Althoff, McGlaun, Spahn, and Lang, 2001] who found out that experts tend to use haptic devices (Mouse, Touchscreen in combination with Speech), whereas normal computer users and beginners prefer combinations of advanced input devices (speech in combination with hand gestures).

[Naumann, 2009] compared multimodality between older and younger users. He shows that older users rated motion interaction worse than younger users and that older users showed a generally lower performance. Overall he states that motion control was the only modality not suitable for older users and that prior knowledge influences the choice of modality.

[Wasinger and Krüger, 2006] examined the role of gender when choosing multimodality: women feel less comfortable using speech in a public environment than men do. [Schüssel, Honold, and Weber, 2013] confirmed this result in their Wizard of Oz study with 53 subjects to study the influencing factors on multimodal interaction using speech, gestures, touch and arbitrary multimodal combinations. Females almost only use the touch screen, males are much more diverse.

In 1994 [Sturman and Zeltzer, 1994] investigated Glove-base input for different applications and systems. Sturman's conclusion is that glove-based input or, more generally, whole-hand input, requires the user to wear a special device which is against the goal of deviceless natural computer interaction. Until now not much has changed: to get accurate hand tracking data, there's still a data glove required nonetheless, the development of 3D sensors like the LeapMotion or Kinect 2 sensor.

[Cohen, Johnston, McGee, and Oviatt, 1998] conducted a user study comparing a graphical user interface with a pen/voice multimodal interface for the task of military force laydown. Using this interface the participant reached a speed improvement of 3.2 to 8.7-fold. Cohen states that multimodal interaction leads to substantial speed an efficiency advantages of multimodal interaction over direct manipulation-based graphical user interfaces.

# 3  Input & Output Modalities

Current human-computer interaction (HCI) is based on the 35-year-old Windows-Icons-Menu-Pointer (WIMP) paradigm. Recent increase in computing power and development of new sensors led to new possibilities for more intuitive user interfaces and input modalities, that should be used in modern systems especially for robots.
Multimodality is defined by [Chatty, 1994] as "a System that features multiple input devices (multi-sensor interaction) or multiple interpretations of input issued through a single device". [Nigay and Joëlle Coutaz, 1993a] uses a quite similar definition of multimodality: "it is the capacity of the system to communicate with a user along different types of communication channels and to extract and convey meaning automatically".
This means that a multimodal system uses speech, gestures and other human input channels to allow the user to interact with the system.

This chapter gives an overview over different input and output channels for human-human or human-computer interaction followed by computer input and output modalities. Later chapters are based on the definition of these input and output modalities and use them within the semantic description for tasks and parameter input.

## 3.1  Why multimodality?

The main goal for using multimodal interfaces is to support more flexible, powerfully expressive, and low cognitive load means of human-computer interaction. They are expected to be easier to learn and are preferred by users for many applications. Some of the more notable advantages are [Maybury and Wahlster, 1998; Ruiz et al., 2010]:

**Robustness** Using different input methods at the same time increases the likelihood of correct recognition, resulting in an increase of the quality of communication between the user and the system. For example if using speech and gestures the user says "Take this green object" and points to a specific item, the system can infer the task that it should do from speech, thus knows that it should pick a green object. By using the gesture it can then determine which object the user meant.

**Naturalness** Human-computer interaction is mainly based on the well-established practices of human-human interaction where the user can choose the desired modality and thus resulting in a high degree of naturalness. This increases the communicative bandwidth and therefore leads to a higher expressivity.

**Flexibility & Efficiency** Flexibility allows individual users to perceive and structure the communication in diverse ways for specific contexts and allows the user to select the modality which is in his opinion the most suitable for the current task. Users can also achieve faster error-correction when using multimodal interfaces.

**Perceptability** Especially modalities with spatial context, like gestures or pen selection, increases the perceptability for the user and gives him a better understanding on how the task will be executed.

## 3.2 Overview

In human-computer interaction there are two main parties involved: a human which uses the system and the computer itself which interprets the input from the human. The following section gives an overview which input channels a human body provides. The next section describes possible sensors which can process these input channels and convert them to the appropriate data format for the computer system.

### 3.2.1 Human Input Channels

The human body has different receptor cells with which a human can perceive the world. Table 3.1 lists the seven main input modalities of the human body, giving a brief introduction on how we interact with other humans.

Obviously not all of these channels are of the same interest for human-computer interaction, especially in the field of intuitive industrial robot programming. The senses of taste and smell might not be very interesting for HCI. This is not due to the fact that there aren't any corresponding output devices available yet, but due to the fact that it's not a very useful channel. One might think of smoke detection to automatically stop a robot if there's smoke, but for a simple smoke detector you don't need such a complex sensor or chemical analysis as the human nose, you can take a vision system and measure the cleanness of the air.

Proprioception on the other side is much more important. It's the self awareness (Kinesthesis) of the body which can be divided into three submodalities: position sense (po-sition of the body including all joint angles), sense of movement (direction, duration and speed), and force sense (applied force through the muscles, or estimating weight by hand).

| Sensory system | Modality | Stimulus energy |
|---|---|---|
| Visual | Vision | Light |
| Auditory | Hearing | Sound |
| Vestibular | Balance | Gravity |
| Somatosensory | Somatic senses: | |
| | Touch | Pressure |
| | Proprioception | Displacement |
| | Temperature sense | Thermal |
| | Pain | Chemical, thermal, mechanical |
| | Itch | Chemical |
| Gustatory | Taste | Chemical |
| Olfactory | Smell | Chemical |

**Table 3.1:** *The seven main input channels of a human body. These different sensory modalities are not processed in isolation. Multimodal areas exist in cortical and sub-cortical areas, such as the posterior parietal cortex and the superior culliculus. Source [Kandel, Schwartz, and Jessell, 2000; Schomaker, 2001]*

## 3.2.2 Human Output Channels

Human output channels are output modalities which we use to interact with the world or change the world state. The output channels are of big interest for human-computer interaction because they define how a user can input data into the system, control it or change the world state.
The human body has four main output modalities:

**Speech** is the most expressive output modality. Using words we can explain complex relations and facts to other people, we can express feelings and describe tasks. Unfortunately it is also the most difficult task for a computer system to elaborate speech data and interpret it (see Section 7.3).

**Force & Movement** Using the muscles, a human body can apply forces to an object for gripping, holding and deforming it. This force may also cause objects to move in space in an arbitrary direction. If an object has appropriate sensors it can sense its pose in the world and infer additional information. Applying force to a surface is an important output modality used in current touch input devices, which spread rapidly. The combination of force and movement allows us for example to write or draw on paper.

**Gesture & Appearance**  The human gesture is another expressive output modality which can indicate a position by pointing somewhere, give commands or even communicate whole sentences (sign language). This also includes the appearance of the human body, e.g. is the human standing, sitting or looking at a specific spot.

**Air**  Our lungs can produce in average 2 PSI (0.14 Bar) air pressure which can be used to blow away small parts or remove dust.

### 3.2.3  Computer Input Modalities

Human output channels produce physical energy patterns varying over time: potential, kinetic and electrophysiological energy. The resulting force, movement and air pressure signals need to be transformed to use them as an input for computer systems. There's a various number of transducer devices available; the following list gives an overview over the most important ones.

#### Keyboard & Mouse

Keyboard and Mouse are by far the most widely used input modalities for computers. In the 1870s the first teleprinter devices were used to simultaneously type and transmit stock market data across telegraph lines. Since the first computers were developed (ENIAC, 1946), keyboards are the main means of data entry. In the year 1984 the mouse was introduced as a consumer device and since then mainly used in combination with graphical user interfaces.[1]
Due to the high circulation level of Keyboard and Mouse, they are well known to most of the users and don't need additional training time. Therefore they are also known for high performance and precision when used as input devices but also for their un-intuitiveness. This input modality can be used for different input operations [Schomaker, 2001]:

- Text input: naming objects and writing text which is displayed directly

---

[1]http://en.wikipedia.org/wiki/Computer_keyboard

- Actions: special keys like $\boxed{\text{Enter}}$ or $\boxed{\text{Esc}}$ generally have the function of accepting or canceling an action. The mouse can be used to select items or change their position (dragging).

- Navigation: using the cursor keys, the user can navigate through items in the user interface. The mouse is a more intuitive way for selecting an item and navigating through different user interface levels.

## Speech Input

In human-human interaction speech is the most important modality for communication and describing tasks. A human can even explain complex relations and situations to another human being. Therefore it's a good idea to use it as an intuitive input modality for computer systems.

Bell Laboratories designed in 1952 the *Audrey* system which could only understand digits spoken by a single voice. Ten years later, IBM demonstrated at the World's Fair its *Shoebox* machine, which could understand 16 words spoken in English. Carnegie Mellon's "Harpy" speech recognition system could understand 1011 words (approximately the vocabulary of an average three-year-old) and was the result of the Department of Defense DARPA Speech Understanding Research program from 1971 to 1976. In the 1980s, the development of Hidden Markov Models was an important breakthrough for speech recognition. Using Hidden Markov Models (HMMs) unknown sounds could be combined to meaningful words. In the early stages of these systems the user needed to dictate the sentences rather than fluently saying a sentence. In 1997, the software *Dragon NaturallySpeaking* was released at what you could speak naturally at about 100 words per minute. Until near the end of the 2010 decade the technology seemed to be stalled at 80 percent accuracy. Then Google released Google Voice Search, an application for mobile phones to use speech as an input method for search queries. In combination with its huge cloud infrastructure they were able to get much better results, now incorporating more than 230 billion words from actual user queries. [Pinola, 2011]

## Touch

Most modern mobile devices and even desktop monitors nowadays have a touchscreen, on which the user can control the system through simple or multi-touch gestures. Rather than using keyboard and mouse, touch input enables the user to directly interact with what is displayed.

The first finger-driven touchscreen was invented by E.A. Johnson in 1965, described in his article [Johnson, 1965]. His invention was a predecessor of current capacitive touchscreens and could only process one touch at a time with only binary information indicating if touched or not. Pressure sensitivity would arrive much later.

In 1970 the American inventor G. Samuel Hurst developed the first resistive touchscreen using a conductive cover sheet and another one that contained the X- and Y-axis. Pressure on the cover sheet caused voltage to flow between the X and Y wires. Resistive touchscreens tend to be very affordable to produce and are durable even in harsh environments. One of the most widely used touch-capable devices at this time was the *PLATO IV* terminal from the University of Illinois. It used infrared emitters and detectors that could sense where the user's finger came down on the screen. At the beginning of the 1980s, Hewlett-Packard began heavily commercializing touchscreens with their HP-150 personal computer which also used infrared sensors. In 1993, IBM and BellSouth launched the Simon Personal Communicator, one of the first cellphones with touchscreen technology. At the end of the 1990s, Wayne Westerman, a graduate student at the University of Delaware introduced the idea of multitouch capacitive displays as we know them today in his doctoral dissertation [Westerman, 1999]. In the following years more and more companies integrated touchscreens into their products. One of those products was Microsoft's Surface 30-inch tabletop which used cameras mounted inside the table looking upwards to the surface detecting finger and hand touches. The final breakthrough for touchscreens is marked by the release of the first iPhone from Apple in June 2007. The combination of a capacitive multi-touchscreen with an intuitive concept of operation was a novelty at this time and therefore a huge success. [Ion, 2013]

### 3D input devices

The first 3D input devices were developed even before the invention of computers. In 1926 C. B. Mirick invented an electrical two-axis joystick used for a remote controlled aircraft. In 1944 During the second world war, German engineers developed a two-axis joystick used to guide missiles against maritime and other targets. In the 1960s the use of joysticks became widespread in radio-controlled model aircrafts by Phill Kraft. Kraft Systems became an important OEM supplier of joysticks to the computer industry [2].

Until now, different devices for 3D input have been developed, which are based on different technologies. Following list gives an overview over the most widely used techniques.

**Joystick**  A joystick consists of a stick that pivots on a base, and buttons mounted on the stick and base for additional input. It is mainly used in aviation and to control video games. Most of the joysticks have three to four degrees of freedom (pitch, yaw, roll and sometimes up/down).

---

[2]http://en.wikipedia.org/wiki/Joystick

**Wired glove**   The first wired glove, the Sayre Glove, was invented in 1977 used to monitor hand movements. It was equipped with flexible tubes, a light source and photocells at the other end. This glove could measure finger and hand bending [Sturman and Zeltzer, 1994]. Nowadays wired gloves use different techniques to detect bending and global positions, like the *5th Glove* or image based data gloves proposed by [Pamplona and Fernandes, 2008] using visual marker at each finger tip and a camera to track movements.

**Spaceball**   The spaceball, also known as SpaceMouse or SpaceNavigator is a 3D motion controller mainly used for manipulating and navigating computer-generated 3D imagery. Its main component is a ball shaped control element which can be moved in 6 degrees of freedom.

### Position tracking

A tracking system determines the position and orientation of the user. Mechanical tracker systems physically connect the user to a point of reference and measure the relative position change. Such trackers are quick and accurate but severely restrict the user's range of motion. Mechanical tracker systems are mainly used for medical surgeries.
For optical tracking, different systems are available which use a camera only setup in combination with generative (model based) and discriminative (classification, regression) algorithms [Emídio de Campos, 2006]. [Gaschler, 2011] uses special visual markers which reflect infrared light emitted by LEDs and recorded by a stereo vision system. The user can hold a tool with buttons in his hand and the system tracks its six degrees of freedom pose. The buttons are used for clicking or selecting objects.
Inertial trackers use gravity, inertia and the earth's magnetic pole to track the orientation. These devices don't require any additional hardware and are very compact, however suffer from orientation drift and cumulative errors.

### Gestures

Gesture recognition enables humans to interact with machines in a natural way without using any additional mechanical devices. The first devices used for gesture recognition were wired gloves (see previous paragraph). Current research mainly focuses on the usage of cameras where the user doesn't need any special glove. Depth-aware cameras use structured light or time-of-flight to generate a depth map from which a specific gesture can be recognized [Liu and Jia, 2004].

The leap motion sensor [3] is even more accurate than cameras: it can track the movement of all ten of the user's fingers. It uses three separate infrared LED emitters and two infrared cameras and achieves an accuracy of about 0.5 to 0.01mm. The only disadvantage compared to camera systems is its rather limited sensory space (500mm x 400mm x 500mm) [Guna, Jakus, Pogačnik, Tomažič, and Sodnik, 2014].

**Gaze**

The user's line of sight can be used as an additional input information for computer systems. It gives an indication at which location the user's attention is focused. Using this data, a system can show additional properties of the object the user looks at, and hide information that isn't of current interest.

There are a number of methods for measuring eye movement, the most popular variant uses video images from which the eye position is extracted. Other methods use for example Electrooculography (measuring the electric potential difference) or optical tracking (infrared light used to detect eye rotations).

**Pen input**

For direct or indirect input in two dimensions there are special pens available. It's a mechanical intermediary especially suited for precise control and selection of small objects. Direct input means that the user can select or interact directly on the screen, indirect input uses a special surface which detects the pen movements. A wide variety of techniques exists. Using resistive, pressure sensitive, capacitive, or electromagnetical transducer technology, the position of a pencil can be accurately measured. Such a pen can provide different information like the position (x,y, velocity, acceleration), pen force, height above the plane, angles and additional hardware button status.

## 3.2.4 Computer Output Modalities

There are many ways how machines can communicate with the user which appeal to different input channels of the human body. This section describes the most important output channels: visual, acoustical and tactile/haptic output.

---

[3]https://www.leapmotion.com/

**Visual output**

Visual output is by far the most commonly used output channel for computer systems and other technical devices. There's a wide variety of devices which produce an image or a signal stimulating the visual sense of a human:

**Monitor** Computer displays are the predominant output channel for technical systems. Early electronic computers were fitted with a panel of light bulbs indicating the states of the registers inside the computer (Zuse Z3 Terminal, 1941). Cathode-ray tubes (CRTs) were first used for memory and it didn't take long before they were used as primitive graphical displays which could show vector lines only (5WAC Console, 1950). In the early 1960s, CRTs were more and more used for displaying text (Uniscope 300, 1964). At this time another display technology emerged: the liquid crystal display (LCD) mainly used for pocket calculators in the late 1970s (Sharp PC-1211, 1980). In the early 1980s, the first color displays were build into computer systems (IBM 5153 CGA, 1983). Throughout the 1980s and 1990s, LCD technology continued to improve, especially the contrast and color capabilities got better, driven by a market boom in laptop computers (TI Extensa 570CDT, 1990). Around 1997 vendors like ViewSonic, IBM or Apple introduced color LCD desktop monitors which could compete with CRT monitors at a reasonable price. Since then LCD monitors outsold CRT monitors and are now used for a wide variety of systems. A recent trend emphasizes monitors that support 3D visualization, higher resolutions and accurate color representation.[4]

The biggest deficiency of monitors appears in the spatial vision stimuli: objects look differently from different angles of view. The lack of spatial effects has a diminishing effect on the sensation of reality and involvement in the content of the scene. A more advanced aspect of this problem is the lack of full immersion in visual scenes, which is being tried to be solved with augmented reality devices.

**Video projector** Video projectors receive a video signal and project it on a projection screen using a lens system. The first projectors date back to the 1970s using CRT (cathode ray tubes). In the 1980s, the trend to use PCs for presentations led to the development of the Epson VPJ-700. Its successor the ELP-3000 became a tremendous hit in 1994. In the following years different companies successfully reduced the size and increased the image quality and brightness of the projectors.

**Augmented Reality** allows virtual imagery to be seamlessly combined with the real world. Around 1960, Morton Heilig created a simulator called *Sensorama* which produced visual sounds, vibration and smell. The first artificial reality laboratory, the *Videoplace*, was developed in the mid 1970s and surrounded the user with an artificial reality which responded to the users movements and actions.

---

[4]http://www.pcworld.com/article/209224/historic_monitors_slideshow.html

Until recent years most of the research has been focused on the technology for providing the AR experience rather than methods for interacting with the virtual content. In the last few years, especially the interaction got more and more important due to better sensors and increasing computing power. The latest developments aim at wearable computers like the *Google Glass* augmented reality glasses announced in 2013. It is connected via Bluetooth to the Smartphone and Internet and responds to voice commands, frame touch and head movement.

**Printer** Electrical printer are used as a static output method producing images or even objects for permanently visualizing data to the user. In 1953, the first computer printer was constructed to be used with the Univac computer by Remington Rand. Since then, various technologies had been adopted. The *IBM 3800* was the first laser printer introduced in 1975, which printed more than 100 pages per minute. The first inkjet printer appeared in 1976, and was produced by Hewlett-Packard (HP), but it took until 1988 to become a home consumer item due to unsolved problems with clogged print heads. In 1970 the first dot-matrix printer, the LA30 by Digital Equipment Corporation, could print at a speed of 30 characters per second and was quite loud. In the last years 3D printing, also known as additive manufacturing, got more and more popular. It uses additives to form solid 3D objects of virtually any shape from a digital model.

### Acoustic output

Acoustic output is achieved by using loudspeaker connected to a computer or other electronic devices. It produces sound in response to an electrical signal input.

The predecessors of the loudspeaker were horns which didn't use electricity and had a disadvantage that they couldn't amplify the sound very much. The first electric loudspeaker was installed by Johann Philipp Reis in his telephone in 1861. It was capable of reproducing clear tones and muffled speech. Primarily further developments in the field of telecommunication were accompanied by new developments of better loudspeakers. In the first half of the 19th century new techniques allowed better sound quality and a wider range of applications aside from telephones.

Nowadays loudspeaker are built into almost every device and are next to computer displays one of the most important output modalities. They can give acoustic feedback to the user or play different sounds, e.g. for warning or confirmation.

**Haptic output**

Haptic output or feedback is a recent development especially in the field of virtual reality. The word derives from the Greek *haptikos* meaning "being able to come into contact with". It's a generalization of the terms *tactile* and *force* feedback. Tactile, or touch feedback applies to sensations felt by the skin which allows users to feel things like texture of surfaces, temperature and vibration. Force feedback reproduces virtual forces applied to the object within the virtual world like weight, inertia or solid boundaries.

The first commercial tactile device was the *Touch Master* in 1993, the first commercial force feedback device was the PHANTOM arm in 1993. The most popular design on the market is a linkage-based system, which consists of a robotic arm attached to a pen where the arm tracks the position of the pen and is capable of exerting a force to the tip of the pen. Another quite popular design for force feedback is the tension based system, in which the object or glove is connected with a cage through cables and a motor. These motors can then exert force on the cables or increase the friction. [Berkley, 2003]

## 3.2.5 Fusion of modalities

Fusion of input modalities is a powerful and complex way of improving the interaction of the user with a specific system. It's the combination and simultaneous processing of different input modalities and allows a more intuitive way of interaction. A user can give a speech command like "Take that object" and at the same time point to a specific object. Therefore the system has to understand from the speech command that the user wants the system to take something, and by evaluating the gesture, define this something as a specific object.

The combination of modalities has been systematically described at the fusion level with the CASE model [Nigay and Joëlle Coutaz, 1993b] and at the user level with the CARE properties [J Coutaz, Nigay, Salber, and Blandford, 1995]. [Nigay and Joëlle Coutaz, 1993b] define the design space along three dimensions: Levels of Abstraction (e.g. process speech as signal, sequence of phonemes, or interpreted as a meaningful sentence), Use of Modalities (sequential or parallel), and Fusion (Combined, Independent). This design space is then used for creating a system which is able to understand commands like "Put that there". CARE stands for Complementarity, Assignment, Redundancy, and Equivalence that may occur between interaction techniques and models the diversity on the user level of multimodal systems.

The fusion of speech and gesture input and the dynamic usage of those two modalities is evaluated in [Vallee, Burger, and Ertl, 2009]. The final result of the user study states that there exist multiple factors which influence the user's satisfaction and that the fusion of speech and gesture input is robust enough for commanding a service robot.

There's a significant difference between human-human interaction and human-machine interaction (cf. [Dahlbäck et al., 1993]) and therefore it makes sense to think about specific combinations of input modalities which are not that common in human-human interaction but may be quite suitable for human-machine interaction.

# 4 Domains & Tasks

Industrial robots are used in many different production domains and therefore require high flexibility on the usage of different tools. Assembly requires a gripper for pick and place tasks whereas welding needs a special welding gun to perform the required tasks. Not only the robot hardware but also the teaching process for industrial robots needs to provide the mandatory flexibility to handle different domains and various tasks. Especially for Small and Medium-sized Enterprises (SMEs) intuitive programming of industrial robots is essential due to the small lot sizes resulting in short reprogramming cycles.

The first section introduces the concept of skills, tasks and processes describing the hierarchical abstraction of an intuitive teach-in process. The Skills section goes into more detail about which skills are required from the robot or the system. Task Analysis mainly focuses on which information is needed from the user to design a new system and to increase the acceptance rate of such systems. The Domains section gives an overview over the most occurring domains in production processes and lists the input modalities which could be used for a specific task within this domain. The last section, Semantic description, combines the knowledge of previous sections into an abstract semantic description of the tasks, their parameters, input modalities and required system components. This ontology can then be used in cognitive robot cells for automatic reasoning on which input modality can be used and is preferred by the user.

## 4.1 Skill, Task and Process

To hide the complexity of a robot system presented to a non-expert worker for teaching a new work-flow, the skills of a robot and human need to be abstracted. This reduces the workload and makes it easier to adapt an existing program to a new product. Especially for human-robot cooperative working, the actions of a human need to be known to the robot, and therefore inserted in an abstracted manner into the process so the robot can inform the human what it is expected for him to do.

**skill**  A skill is a basic action-block which is performed by the robot and normally requires additional parameters for its execution, e.g. *move to position*, *close gripper*, *open gripper* or actions requiring additional atomic steps like *drill hole* (switch on, move downwards to drill the hole, move out of hole, switch off). Skills are provided by the hardware and are very basic operations available to the user.

**task**  The introduced skills can be combined to reach a specific goal. Such a combination is called a task, e.g. *pick object* combines the skills *move to object* and *close gripper*. A task can have different skill mappings providing the same functionality, thus there may exist multiple skill implementations for the same task that use for example different sensors or algorithms. Since tasks are mainly object based, e.g. pick object, they need a specific knowledge base and sensor system in the background to detect the object and map its position to absolute coordinates reachable by the robot. The knowledge base also defines additional parameters for the skills, which can be automatically inferred from the information given by the user, like the electrical current needed for a welding gun if a specific type of metal should be welded.

A task may also be a specific job which needs to be carried out by a human worker, such as handling flexible parts (cables, labels, gasket). These specific human tasks can be inserted in the process to tell the robot that it needs to wait for the human worker to finish his job.

**process**  A process combines all the tasks needed for the production step and thus it can be seen as a program which is executed in a loop for each product. It is an abstract high-level description of a robot program.

Figure 4.1 shows an assembly process stripped down into tasks and skills. By providing predefined skills and low-level tasks to the programmer, the whole assembly process is easier to program and can be easily adapted to new assembly workpieces.

## 4.2  Skills

A Skill is a basic representation of an action a robot is capable executing. In task-based robot programming, skills can be seen as commands which are called by a task and require different parameters. The skills are provided by the hardware or robot controller and therefore introduce a Hardware Abstraction Layer (HAL) for higher level tasks. The HAL makes it possible to easily switch between different robot systems using the same high level task.

Skills are not only implemented by the robot controller itself, but also by other system components like computer vision or visual output devices. The following list gives some examples for skills which are used in Section 4.4. Each skill requires a set of parameters which are inferred on the task level by an inference engine or knowledge database based on other parameters defined for the task. For some of the skills uncertainties are specified which may arise when implementing or executing the skill.

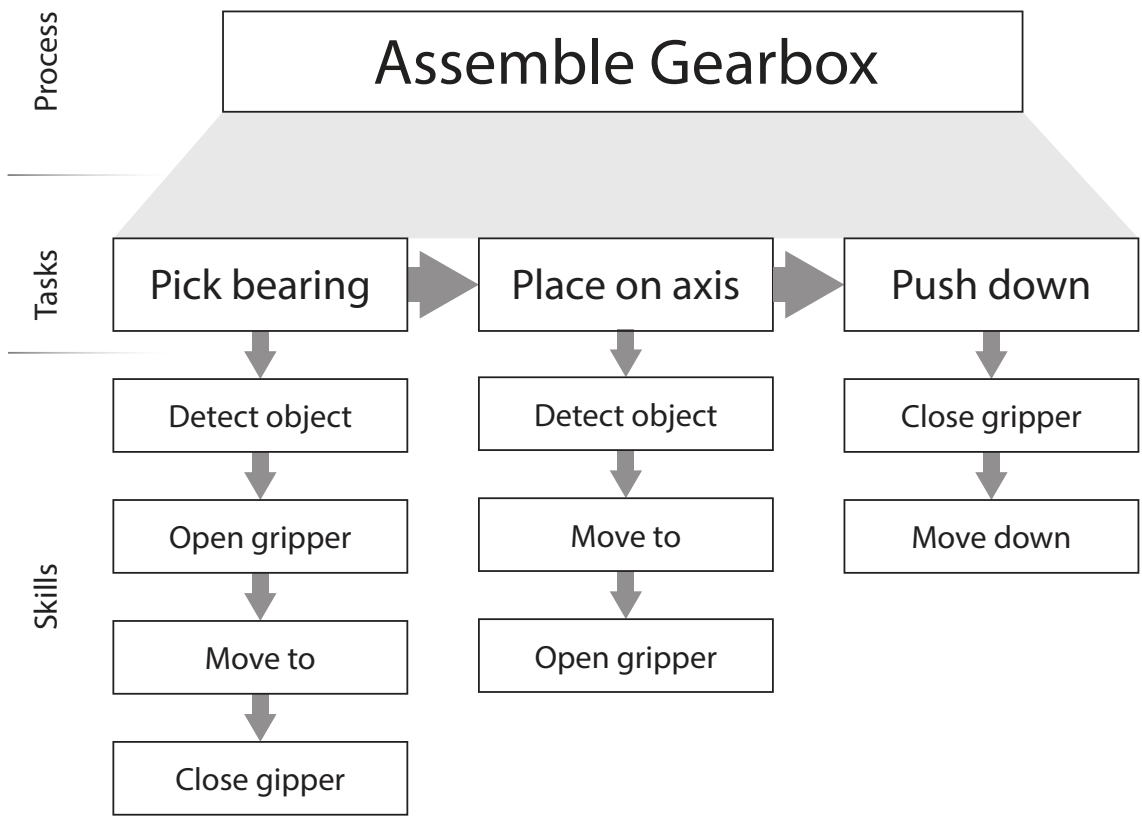

**Figure 4.1:** *Hierarchical abstraction of an assembly process into process, tasks and skills. The process describes parts of an assembly of a gearbox where the bearing is put on the axis and needs to be pushed down until it touches the collar.*

**move to** (`6D pose`, `speed`, `acceleration`): Moves the robot to the specified position using the given speed with given maximum acceleration. The position coordinate should be given as the Tool Center Point (TCP) and the orientation based on the current tool attached to the robot arm.
*Uncertainties*: What is the base coordinate system (e.g. robot base, word base)? Is it the Tool Center Point (TCP) coordinate? Is the TCP correctly set up? What's the initial orientation of the tool on which the given angles are based? When using a 7-DOF robot arm, does the 7th DOF matter for the current task or not?

**dual arm move to** (`6D Pose`, `6D Pose`, `speed`, `acceleration`): Moves two robot arms simultaneously to the specified Tool Center Point (TCP) coordinates and given orientation.

*Uncertainties*: Which parameter is for which arm? Do the arms need to move simultaneously or is sequentially/delayed also allowed (e.g. to avoid collisions between the arms)?

**detect object** (`3D object model`): Detects the position and pose of the object which looks similar to the given 3D object model within the workplace.
*Uncertainties*: If there are multiple objects of the same type detected, which position should be returned? What is the orientation of a symmetric object (e.g. circular object or sphere)?

**open gripper** (): Opens the gripper to release the object currently grabbed or to grab a new object.
*Uncertainties*: How to handle the situation if there's currently another tool attached to the robot arm (does the robot have an automatic tool changer)?

**close gripper** (`force`): Closes the gripper until the specified force is reached. Used to grab objects.
*Uncertainties*: Is the gripper attached to the robot?

**set welding current** (`welding current`): Sets the electrical current for the welding gun. The current is dependent on the material which is being welded.

**set welding speed** (`welding speed`): Sets the moving speed for the welding gun. It is dependent on the material and the type of welding.

**weld point** (`6D Pose`): Weld a single point at given position keeping the welding gun at the given orientation.
*Uncertainties*: Is the welding gun attached to the robot? See uncertainties for *move to* skill.

**weld seam** (`6D Pose array`): Weld a seam along the given points and orientation and interpolates between the points.
*Uncertainties*: Is the welding gun attached to the robot? Which interpolation method should be used between the points? See additional uncertainties for *move to* skill.

**saw along trajectory** (`6D Pose array`): Uses a sawing blade to saw along the given points and orientation and interpolates between the points.
*Uncertainties*: Is the sawing tool attached to the robot? Which interpolation method should be used between the points? Which sawing technique should be used (sawing blade, laser cutting, stitching)? See additional uncertainties for *move to* skill.

**switch on/off grind/deburring/milling tool** (): Switches the tool attached to the robot arm on or off. This tool can be for example a grinding, deburring or milling tool.
*Uncertainties*: Is the correct tool attached to the robot? Is the current tool position safe to switch it on? Should the tool only be switched off if the robot isn't currently moving?

**drill screw** (`depth`, `force`): Drills a screw into the material at current robot arm position stopping at given depth and drilling with given force.
*Uncertainties*: Is the drilling tool attached to the robot? How is the screw fed to the drilling tool? Is the robot at the correct position? How can the robot detect if the screw is set tight?

**apply glue** (`amount`, `6D Pose array`): Applies the given amount of glue moving along the points given as parameter. The connection between the points is interpolated.
*Uncertainties*: Is the gluing tool attached to the robot? Which interpolation method should be used between the points? How can the robot detect a clogged glue outlet? How much time does the robot have to apply the glue until it is dry? See additional uncertainties for *move to* skill.

## 4.3 Task Analysis

To understand the needs of a worker programming an industrial robot and to determine the type of required parameters, a task analysis should be carried out on each task within every domain.
Catherine Courage defines Task Analysis as:

> Task analysis means understanding users' work. Thus, task analysis encompasses all sorts of techniques, including naturalistic observations and interviews, shadowing users or doing "day in the life of" studies, conducting ethnographic interviews, and observing and listening to users who are performing specific tasks. It includes gathering information that leads to insights about users' lives at work or at home, to scenarios and use cases, and sometimes to detailed flowcharts of work processes or specific procedures.
> [Courage, Jain, Redish, and Wixona, 2012]

The most important statement of this definition is that task analysis is "information gathering that leads to insights about users' work" which allows the system designer to fit the new system to the needs of the workers and therefore increases the acceptance of new systems which is especially important for Small and Medium-sized Enterprises (SMEs).

It's also important to understand the user, the task he is performing and the user's environment (physical, technological, cultural, social and political). It may help to answer the following questions [Courage et al., 2012]:

1. Users: Who are they? What do they know about the technology? What do they know about the domain? How motivated are they?

2. Tasks: What are the tasks the user needs to accomplish?

3. Users' environments: Physical situation in which the tasks occur; technology available to the users; social, cultural, language considerations.

The following section will mainly focus on the users' environment for task analysis.

## 4.4 Domains

This section gives an overview over typical application domains for industrial robots, primarily suited for Small and Medium-sized Enterprises (SMEs), which play a major role within the SMErobotics [1] project, namely *Assembly*, *Welding*, *Woodworking* and *Metal processing & Bonding materials*. Each domain is described briefly and contains a list of the characteristic tasks within this domain.

This list of tasks for each domain should give an overview over the most commonly required tasks for the domains and is by far not complete. Each task is kept fairly general to ease the adaption to more complex and in more detail specified tasks. It is briefly explained how the task can be used and which skills are required from the robot. Additionally all the parameters necessary to execute the task, including type and input modalities, are listed.

Especially when using robots to execute the task, there may arise a huge number of uncertainties and problems which a human worker automatically resolves but a robot can't resolve on its own. Therefore each task description also includes a few examples for uncertainties with which the programmer of such a system has to cope with.

---

[1]http://www.smerobotics.org

### 4.4.1 Assembly

Assembly is a manufacturing process in which parts are added subsequently to an assembly until the final product is produced. It is one of the most required production steps within Small and Medium-sized Enterprises (SMEs). When using a robot for assembly tasks, it has to cope with increasing complexity due to growing product quality requirements and to handle small components. The main reason for the high complexity is the difficulty of conventional industrial robots to adjust to any sort of change. Especially for SMEs, having small lot sizes and therefore a high product variety, adaptability and flexibility is important. They can't afford a single robot for one specific production step. Instead SMEs need a robot that can be used for different assembly steps and provides the required accuracy, e.g. for mounting a bearing onto a shaft.

Programming a robot to execute a specific process within the assembly domain involves different types of tasks. The following list gives an overview over the most common tasks within assembly and analyses each task regarding the problems and uncertainties which may arise and lists the required parameters which are needed by the robot to successfully execute the task:

**Pick Object**

The *Pick object* task is defined as moving the robot to an object and closing the gripper to lift that specific object.

**Required skills:** move to, detect object, close gripper

**Parameters:**

- *Object to pick*: The user needs to define which object the robot should pick up.
    - Type: 3D model of the object
    - Input modalities: Point by hand (gesture pointing to the object), Speech (name object or describe position), Touch (select from object library), Pen Input (point with 3D pen to object and click), Keyboard & Mouse (type the name and select from library)

**Uncertainties:** Is the object already on the table or is it palletized. If the user points in between of two objects, which one should be selected? If the same type of object is multiple times on the table, the user has to identify which one to pick (e.g. "Robot pick the left bearing"). If objects are already assembled and the robot should disassemble them, how does the user define if the robot should take the assembly or disassemble the assembly by removing an object.

Does the object require a specific gripping position, that the system shouldn't infer automatically? What happens if the object to pick isn't there anymore?

## Place Object

The *Place object* task tells the robot where to place the previously picked object.

**Required skills:** move to, open gripper, detect object

**Parameters:**

- *Location to place*: Location on the table or any other plane area where the robot should place the object which is currently in it's gripper.
    - Type: 3D coordinate
    - Input modalities: Point by hand (point to a location on the table/object), Speech (describe the location, e.g. center of the table), Touch (coordinate input), Pen Input (point to a location and click), Keyboard & Mouse (coordinate input)

**Uncertainties:** Should the object be placed directly on the table or stacked onto another object which is already on the table? Are intermediate positions required, i.e. must the robot move exactly above the object to move it down straight or is there a specific path to follow, so that the object is maneuvered around obstacles? How does the robot know if it is a non rigid body? How should the robot handle non rigid bodies and avoid collisions due to unknown object dimension?

## Pick & Hold

*Pick & Hold* is a task where the robot picks an object and holds it at a specific pose in front of a human worker so that he can assemble or work with the object.

**Required skills:** move to, open gripper, detect object

**Parameters:**

- *Object to pick*: The object which should be picked up by the robot. See *Pick Object* task on page 35.

- *Hold pose*: Pose of the object where the robot should hold it.

– Type: 6D pose
– Input modalities: Point by hand (use gestures, e.g. draw a rectangle in the air where the object should be), Speech (define the pose by describing it), Touch (use a 3D visualization), Pen Input (like point by hand, draw in the air and define the space), Keyboard & Mouse (define pose by setting coordinates and rotations manually)

**Uncertainties:** How can it be defined that the objects should be rotated after a specific step (e.g. the human worker needs to assemble something on the front and back side of the object)? How tell the robot that the assembly step is finished and it should continue with the next task? How can the robot check if the human executed his work so that the next steps don't fail? (cf. uncertainties for *Pick Object* task on page 35).

## Assemble objects

The *Assemble objects* task is used to assemble two objects based on a defined assembly pose.

**Required skills:** move to, open gripper, close gripper, detect object, dual arm move to

**Parameters:**

- *Object to assemble*: The object which should be picked and assembled with another object.
  See *Pick Object* task on page 35.

- *Assembly*: An already assembled object or a single object with which the previously selected object should be assembled.
  See *Pick Object* task on page 35.

- *Assembly constraints*: How should the objects be put together/assembled.

  – Type: object constraints
  – Input modalities: Speech (define constraints, e.g. "Put bearing on axis and align them concentric"), Touch (3D editor where the user can define constraints between objects), Keyboard & Mouse (define constraints similar to a programming language)

**Uncertainties** : Does the assembly step require a specific trajectory/path or can it be inferred automatically by the robot? How can the user define a trajectory if the system doesn't find any? Does the assembly step require a human worker (especially for non rigid assembly objects like cables or gaskets)? How does the robot know how the objects should be put together (just push or is a rotation needed, like screwing)? If the objects are hold together with clips, how does the robot get feedback that the clip is latched (human workers normally listen for the "click" sound)? How much force is needed? Can the objects be assembled on the table or is a dual arm assembly station needed? How does the system know if an object should be put for example onto an axis and then pushed to the collar?

## 4.4.2  Welding

In welding, usually metals or thermoplastics are joined by coalescing the materials. There are many different welding methods, requiring various hardware components and techniques. In this section the focus lies on the mainly used *shielded metal arc welding* and *Flux-cored arc welding*.

It's a complex manufacturing step where different parameters must be set correctly (e.g. current, speed, angle) to reach the required product quality. These parameters must be set manually or can sometimes be automatically inferred or read from a knowledge database. Thus programming a robot to weld two materials and bond them together involves different types of tasks. The following list gives an overview over the most common tasks within welding, analyses each task regarding the problems and uncertainties which may arise, and lists the required parameters which are needed by the robot to successfully execute the task:

**Define material**

For welding it is important to know which materials are welded to set the correct welding parameters like current and speed. These parameters can be inferred from a knowledge base or can be predefined values for each material.

**Required skills:**  set welding current, set welding speed

**Parameters:**

- *Material*: The material of the object to weld.
    - Type: material (string or id)
    - Input modalities: Speech (name of the material), Touch (select material from a list), Keyboard & Mouse (write name of the material)

- *Thickness*: Thickness of the material to be welded.

    - Type: number (unit millimeters)
    - Input modalities: Speech (say the number), Touch (select a thickness using a slider), Keyboard & Mouse (write the number into a text field)

**Uncertainties:** Can the system automatically infer the material based on its knowledge about the object? How to handle two different materials which need to be welded together? If the material has varying thickness, how to define it and how should the robot cope with it?

## Point welding

Point welding is a process where two materials, mostly plates or flat parts, are bonded together by one or multiple welding points.

**Required skills:** move to, weld point

**Parameters:**

- *Object to weld*: The object where point welding should be applied.
    See *Pick Object* task on page 35.

- *Position*: Single point or vertex where the welding gun should apply current to weld the parts together.

    - Type: 3D point or vertex
    - Input modalities: Point by hand (point on an object already lying on the table), Speech (describe the point), Touch (use a touch-friendly 3D visualization), Pen Input (use the pen to select a position), Keyboard & Mouse (3D visualization or manual coordinate input)

- *Material*: Material of the object to weld.
    See previous *Define material* task.

**Uncertainties:** How can the robot check if the two parts are correctly aligned before welding them together? Is there a fixation or another robot arm required to hold the parts at a specific position, or are they already assembled in a previous step? If the welding point isn't reachable, how should the robot behave? What's the best orientation of the welding gun for optimal results?

**Seam welding**

With seam welding, two materials or parts of an assembly are connected along an edge or seam. These parts are often aligned perpendicular and therefore there is a well defined edge between those two.

**Required skills:** move to, weld seam

**Parameters:**

- *Object to weld*: The object where seam welding should be applied.
  See *Pick Object* task on page 35.

- *Edge*: Edge defined of multiple points or an edge on the object, which the robot should follow for seam welding.

  - Type: List of 3D points or an edge
  - Input modalities: Point by hand (point along an edge on an object already lying on the table), Speech (describe the edge), Touch (use a touch-friendly 3D visualization), Pen Input (use the pen to select one or multiple connected edges), Keyboard & Mouse (3D visualization or manual coordinate input)

- *Material*: Material of the object to weld.
  See previous *Define material* task.

**Uncertainties:** How can the robot weld along twisted planes or edges? Does the welding require special steps like filling a gap or seam (move over the same edge multiple times)?
See also previous *Point welding* task.

## 4.4.3 Woodworking

Woodworking or wood processing describes manufacturing and machining of work-pieces made of wood, some of which are sold directly as goods and some as components assembled with other materials into functional products.

Especially in the field of prefabricate houses, robots are used to handle the heavy objects and provide accurate machining of the workpieces. Since each house is different from previously produced ones, this field of application for robotics requires a high flexibility and easy adaption to changes.

The following list gives an overview over the most required and used tasks for woodworking whereas this domain also requires different tasks from the Assembly domain in Section 4.4.1. Each task is analyzed regarding the problems and uncertainties which may arise and the required parameters which are needed by the robot to successfully execute the task:

**Sawing**

Sawing is one of the main tasks within the woodworking domain. It's used to cut out parts of a wood panel or to trim planks and bigger panels. It requires a specially equipped robot using a saw blade to cut through the material.

**Required skills:** saw along trajectory

**Parameters:**

- *Object to saw*: The object where the robot needs to cut out parts or trim. See *Pick Object* task on page 35.

- *Start position*: position where the trajectory should start. This position normally needs a drilled hole to insert the saw blade if the trajectory is a closed loop.

    – Type: 3D or 6D pose
    – Input modalities: Point by hand (point on an object already lying on the table), Speech (describe the point), Touch (use a touch-friendly 3D visualization), Pen Input (use the pen to select a position), Keyboard & Mouse (3D visualization or manual coordinate input)

- *Trajectory*: The path or trajectory along which the sawing blade should move to cut through the material or wood.

    – Type: list of 3D or 6D points/poses
    – Input modalities: Point by hand (point along the path on an object), Speech (describe the path, e.g. circle at the center with a diameter of 30cm), Touch (use a touch-friendly 3D visualization), Pen Input (use the pen to move along the path, like drawing on the wood), Keyboard & Mouse (3D visualization or manual coordinate input)

**Uncertainties:** Is the hole in the wood to insert the sawing blade already there? Is another robot arm needed to hold the cut off part to avoid squeezing of the blade? Should the robot use other cutting technologies like laser cutting or stitch sawing?

**Grinding**

The grinding task describes the process of reducing unevenness on wood or other materials by using a rotating grinding paper. It's often used to make a surface smoother or to reduce the thickness by a small amount.

**Required skills:** move along, grind on, grind off

**Parameters:**

- *Trajectory*: The path or trajectory along which the robot should grind.
  - Type: list of 3D or 6D points/poses
  - Input modalities: Point by hand (point along the path on an object), Speech (describe the path, e.g. circle at the center with a diameter of 30cm), Touch (use a touch-friendly 3D visualization), Pen Input (use the pen to move along the path, like drawing on the wood), Keyboard & Mouse (3D visualization or manual coordinate input)

- *Grinding depth*: Defines how much material the robot should grind off.
  - Type: number (unit millimeters)
  - Input modalities: Speech (say the number), Touch (select a depth using a slider), Keyboard & Mouse (write the number into a text field)

- *Coarseness*: number defining the coarseness of the grinding paper to use.
  - Type: list of available grinding papers.
  - Input modalities: Speech (say the number), Touch (select a coarseness using a slider or select from a list), Keyboard & Mouse (write the number into a text field)

**Uncertainties:** If a plane area should be grinded, how can this be defined? How can the robot handle twisted planes? How can the robot exchange the grinding paper if a different coarseness is needed? How much force does the robot need to apply for each coarseness? In which direction should the robot move (when grinding wood, it's important to consider the figure of the wood)?

**Deburring/Rounding corner**

The *Deburring* or *Rounding corner* task is used to remove sharp edges of an object.

**Required skills:** move along, switch deburring tool on, switch deburring tool off

**Parameters:**

- *Deburring type*: deburring can be achieved either by grinding or cutting of the edge in an angle of 45 degree or by grinding the edge in a rounded shape.

    – Type: list (values: rounded, flat)
    – Input modalities: Speech (say/describe the deburring type), Touch (select from a list), Keyboard & Mouse (select from a list)

- *Radius*: radius for rounded deburring or depth for flat deburring

    – Type: number (unit millimeters)
    – Input modalities: Speech (say the number), Touch (select a depth using a slider), Keyboard & Mouse (write the number into a text field)

- *Edge*: Edge defined by multiple points or an edge on the object which the robot should deburr.

    – Type: List of 3D points or an edge
    – Input modalities: Point by hand (point along an edge on an object already lying on the table), Speech (describe the edge), Touch (use a touch-friendly 3D visualization), Pen Input (use the pen to select one or multiple connected edges), Keyboard & Mouse (3D visualization or manual coordinate input)

**Uncertainties:** Which deburring tool should be used (drill, grinding paper)? How handle twisted or peaked edges? What's the optimal angle for the deburring tool?

### 4.4.4 Metal processing & Bonding materials

Machining in general describes a process in which a piece of raw material is processed until it reaches a desired final shape. This process consists commonly of a controlled material removal, also known as subtractive manufacturing, and is one of the main tasks within metal processing, but also for processing wood or other materials.

Bonding materials, in contrary, can be seen as material adding, since it connects two materials so that they are fixed in a specific relative position.

The following list contains two example tasks for metal processing, especially for machining, and one example task for material bonding using adhesives, including the required parameters and possible uncertainties and problems. Some of the tasks of the Woodworking domain in Section 4.4.3 can also be adapted to the metal processing domain and are therefore not listed within this section.

### Milling

Milling is a task where the robot removes or cuts out a special amount of material from the workpiece and is mainly used for creating recesses.

**Required skills:** move to, milling tool on, milling tool off

**Parameters:**

- *Object to mill*: The object where the robot needs to mill material off.
  See *Pick Object* task on page 35.

- *Milling depth*: Defines how much material the robot should mill off.
  - Type: number (unit millimeters)
  - Input modalities: Speech (say the number), Touch (select a depth using a slider), Keyboard & Mouse (write the number into a text field)

- *Trajectory*: The path, trajectory or area along which the milling tool should move to mill out the material.
  - Type: list of 3D or 6D points/poses or an arbitrary polygon
  - Input modalities: Point by hand (point along the path on an object), Speech (describe the path, e.g. circle at the center with a diameter of 30cm), Touch (use a touch-friendly 3D visualization), Pen Input (use the pen to move along the path or define the area), Keyboard & Mouse (3D visualization or manual coordinate input)

**Uncertainties:** Which milling tool should be used? How fast should the milling head move? If an area is selected, how should the robot move in this area to mill off with an optimal result?

### Screw

Screwing is used to fixate two or more objects so that the movement between these objects is minimized. It requires holes and the corresponding screws to hold them together.

**Required skills:** move to, drill screw

**Parameters:**

- *Screw type*: screw type to use. It can be selected from a list of available screws.

  – Type: list
  – Input modalities: Speech (say/describe the screw), Touch (select from a list), Keyboard & Mouse (select from a list)

- *Objects to screw*: Objects which should be screwed together.
  See *Pick Object* task on page 35.

- *Hole*: Hole on the objects through which the screw should be rotated.

  – Type:
  – Input modalities: Point by hand (select the holes on the objects), Speech (describe the hole, e.g. third hole from bottom), Touch (use a touch-friendly 3D visualization), Pen Input (use the pen to select the holes), Keyboard & Mouse (3D visualization or manual coordinate input)

**Uncertainties:** Is the hole for the screw already drilled? If a hole was drilled in a previous step, the selection of the hole for screwing needs a simulated/modified model where the hole exists. How is the screw put into the hole to start screwing? How much rotational force is required? How much pushing force is required? How can the robot check if the objects are correctly aligned before screwing?

## Adhesive bonding

Adhesive bonding or gluing is used to connect different materials and fixate their relative position to each other.

**Required skills:** move to, apply glue

**Parameters:**

- *Glue type*: glue type to use. It can be selected from a list of available glues and should be dependent on the material to glue together.

  – Type: list
  – Input modalities: Speech (say/describe the glue), Touch (select from a list), Keyboard & Mouse (select from a list)

- *Trajectory*: The path, trajectory or area along which the glue should be applied.
    - Type: list of 3D or 6D points/poses
    - Input modalities: Point by hand (point along the path on an object), Speech (describe the path), Touch (use a touch-friendly 3D visualization), Pen Input (use the pen to move along the path), Keyboard & Mouse (3D visualization or manual coordinate input)

- *Amount of glue*: Amount of glue to apply when moving along the trajectory.
    - Type: number (unit milliliters or grams)
    - Input modalities: Speech (say the number), Touch (select an amount using a slider), Keyboard & Mouse (write the number into a text field)

**Uncertainties:** Can the amount of glue be automatically inferred? Which pattern should be used to apply glue to a bigger area (e.g. S-shape or just lines)? How are the parts put together after applying the glue? How much time remains to put the parts together?

## 4.5 Semantic description

A semantic description or ontology describes relations between different objects and is used to formally represent knowledge from a specific domain. Based on the Task Analysis from previous sections within this chapter, an ontology was created, that represents the hierarchical structure of Process, Tasks, Skills and shows the relations of different system components like sensors with for example software components and task parameters.

This ontology can then be used to automatically infer, if a specific input modality is available to the system, and which input modality is the preferred one for each task parameter.

The top level ontology is shown in Figure 4.2. Figure 4.3 lists all the available tasks within the ontology, Figure 4.4 shows a subset of skills that are mapped to tasks. A task can have different skill mappings providing the same functionality. Figure 4.5 lists all parameters which may be required by a task. Figure 4.6 shows the relations for the `ObjectModel` parameter: it shows all the parameters which require an object model, shows its data type and the preferred input modality. The object model parameter type is also used by the `DetectObject` skill, shown in Figure 4.7. The ontology additionally includes a representation of all input and output modalities (Figure 4.8) as well as the main hardware and software components (Figure 4.9).

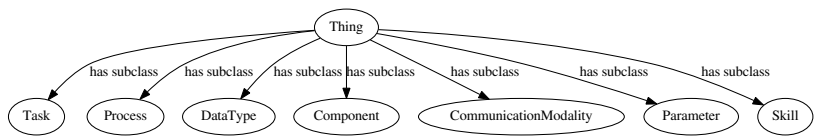

**Figure 4.2:** *The top level hierarchy contains Processes, Tasks and Skills, Parameters for each Task, data types of the parameter, hardware and software system components like sensors or specific software algorithms for example speech recognition and communication modalities for input and output.*

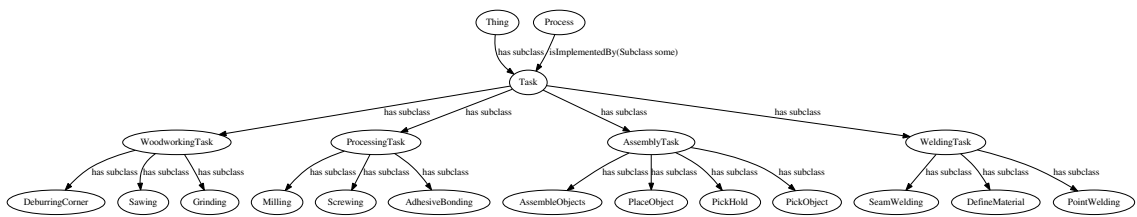

**Figure 4.3:** *A process is implemented by one or more tasks. The tasks can be separated into different groups: woodworking, processing, assembly and welding.*

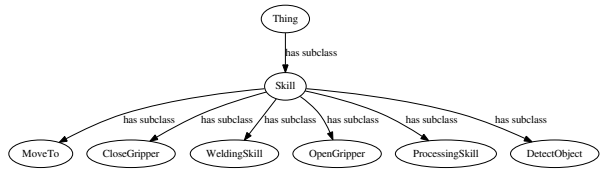

**Figure 4.4:** *A skill is used by tasks and represents the lowest hierarchical representation of a process. MoveTo, CloseGripper, OpenGripper and DetectObject are common skills. There are additional skills for each domain grouped into WeldingSkill and ProcessingSkill.*

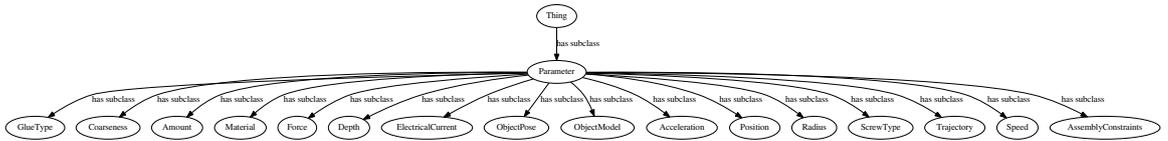

**Figure 4.5:** *Each Task or Skill requires specific parameters for successful execution. This figure lists all the parameters used within the ontology.*

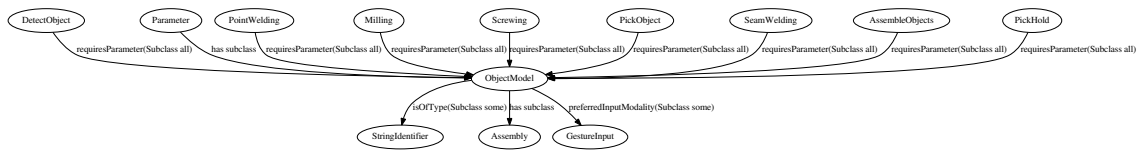

**Figure 4.6:** *This Figure shows an example partial ontology for the ObjectModel parameter. It is a subclass of* Parameter *and is required by different skills (DetectObject) and tasks (PointWelding, Milling, Screwing, PickObject, SeamWelding, AssembleObjects, PickHold). It is of the data type* `StringIdentifier`*, and has a subclass Assembly (which is a special object model) and the preferred input modality for setting an object model is gesture input. This ontology is used for inferring data types and deciding which input channel should be proposed by the system.*

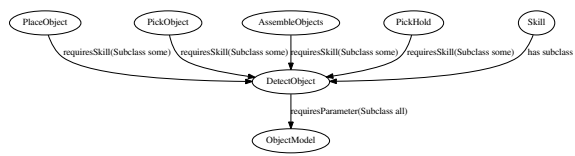

**Figure 4.7:** *Similar to previous Figure, this one shows a part of the ontology which represents the* `DetectObject` *skill. It's required by different tasks and requires an ObjectModel parameter (see Figure 4.6).*

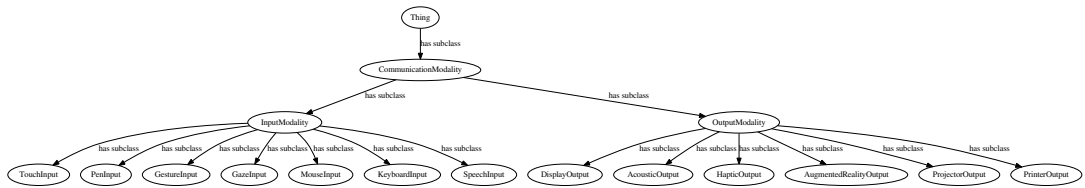

**Figure 4.8:** *Communication modalities are used to interact with the system. They are divided into two groups: input and output modalities and referenced by parameters and data types.*

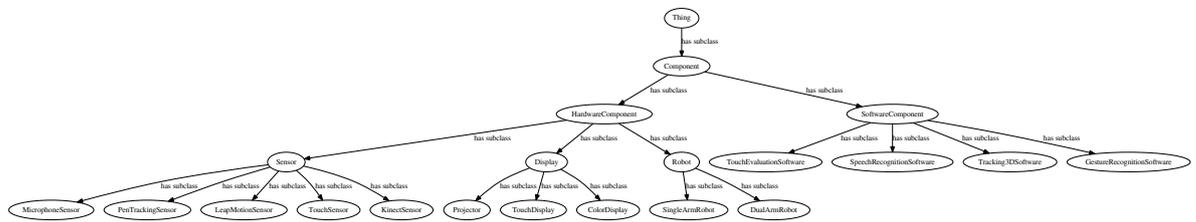

**Figure 4.9:** *To know if a specific input or output modality is available, the system has to know which hardware and software components are built into the robot cell. The component ontology represents those components (e.g. Sensors, Display, Robot or available software).*

# 5 Graphical User Interface

Each and every system requires some kind of visualization or visual feedback for the user so that he knows what the system is currently doing and which data is required from the human operator. It's common ground to use a graphical user interface presented on a display to provide the feedback and give a way to interact with the system. For this master thesis a graphical user interface (GUI) was developed with the main goal to offer an intuitive touch friendly user interface for task-based industrial robot programming, and to provide a visual feedback component for current performed actions of the robot. Flexibility and portability (multi device support) was a main goal for this user interface. It is based on the abstract semantic description introduced in the previous chapter and used for the user study (described in the next chapter) that is one of the main parts of this thesis.

## 5.1 Frameworks & Technologies

The first step for the development of the graphical user interface was to evaluate different frameworks and technologies to build on to. The main requirement for the user interface was flexibility and portability. Flexibility is important to allow quick extension and implementation of new system components and functions. The goal of portability is to provide a system which can run on different devices (Windows, Linux, OS X, Android, iOS) with varying screen resolutions.

To get a portable application which runs under different operating systems, web development technologies like HTML5, Javascript and CSS3 are the best candidates. Using a responsive CSS design additionally comes with user friendly visualization on different screen sizes. Using so called Web-Applications, the user can open the application on his smartphone, tablet or workstation PC. This is a main advantage compared to proprietary programmed apps for each operating system. A drawback of Web-Apps is that they constantly need a connection to the webserver delivering the app. Since the application will be used within factory and production halls, this disadvantage shouldn't be a big problem.

Flexibility is reached by using standardized component interfaces, well known programming languages and software engineering practices such as the Model-View-Controller design pattern.

After deciding that the client side should be written as a Web-Application, the next step was to decide which server software to use. After considering different possibilities like PHP[1] and Zend Framework[2] or Perl[3] and Catalyst Framework[4] the decision fell for using Node.js[5] and express[6] on the server side. This has the advantage that the client- and server-side uses the same programming language, namely JavaScript, and it's possible to reuse components for different parts of the software.

On the client side backbone.js[7] and Bootstrap[8] is used. Backbone.js is a JavaScript library which provides a powerful basis for Single-Page Web Applications using the Model-View-Controller design pattern. The Boostrap HTML, CSS and JavaScript framework developed by Twitter is useful for designing responsive Web Applications.

The Less CSS pre-processor[9] is a useful tool for writing well structured and quickly adaptable and manageable CSS code.

The GUI also required some kind of 3D visualization to show object models and to allow interaction with these models. Since the application is running in the browser, is was straightforward to use WebGL for this kind of visualization. There are different JavaScript libraries available, which integrate WebGL and provide easy access to its functions. One of the most advanced and widely used libraries is the three.js library [10] which was used for the implementation within this GUI.

## 5.2 Design

The basic idea for the design was to create a Single-Page Application that avoids loading and processing time for each clicked link and it has the possibility to dynamically add and remove parts from the page based on the user's input.

These dynamically added and removed parts are called cards. Cards are grouped using so called pages. Figure 5.1 shows a first mockup of the GUI explaining the concept of cards and pages.

Opening a new page adds it to the right hand side of the rightmost page currently shown by using a fly-in animation which moves the page from right to left until it reaches its final position. If the number of pages don't fit on the current screen, the leftmost page is removed from the current view by moving it out of the view at the left side and the remaining pages slide over to the left, to fill the free space. Each page's size is dynamically adapted to the required size of each card within this page.

---

[1]http://php.net/
[2]http://www.zend.com/
[3]https://www.perl.org/
[4]http://www.catalystframework.org/
[5]http://nodejs.org/
[6]http://expressjs.com/
[7]http://backbonejs.org/
[8]http://getbootstrap.com/
[9]http://lesscss.org/
[10]http://threejs.org/

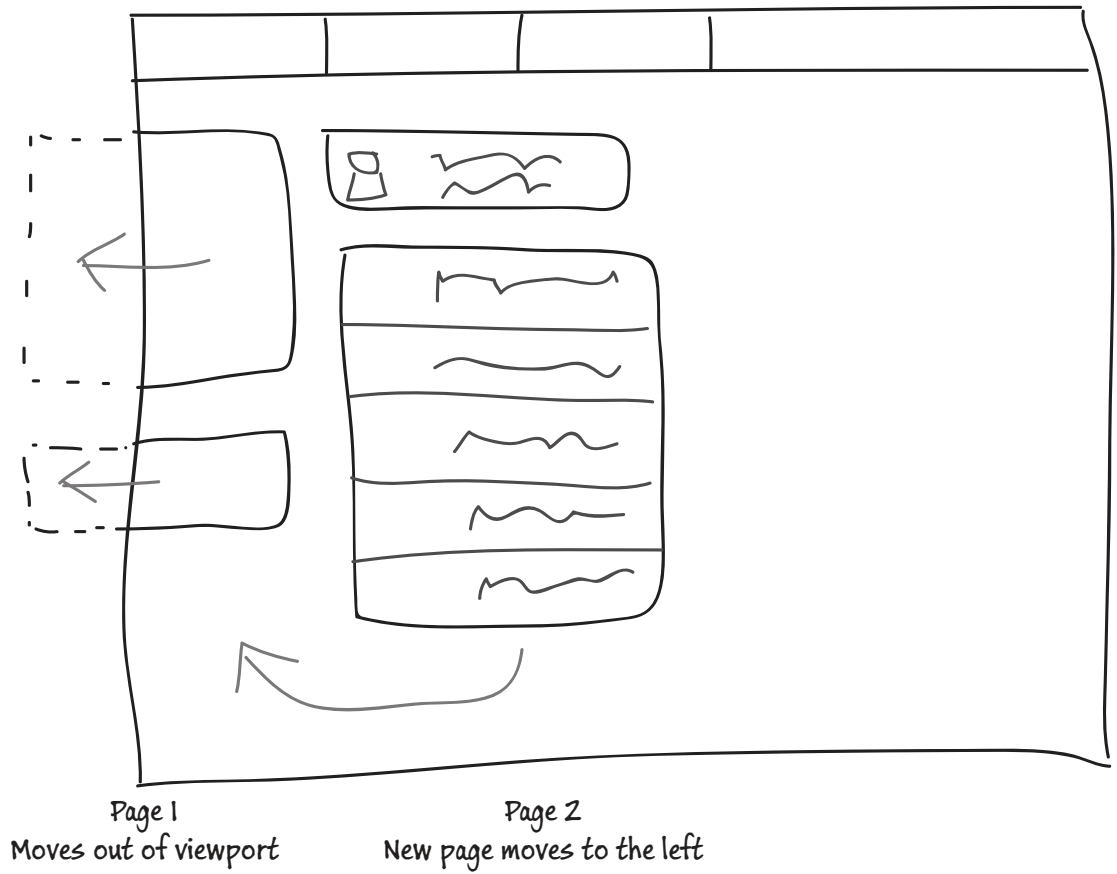

Page 1                    Page 2
Moves out of viewport     New page moves to the left

**Figure 5.1:** *First mockup of the GUI. The top bar contains the main navigation. If a new page gets added and there isn't enough space, the old one is moved out to the left and the new page fills the new available space. A page contains cards (boxes) which are containers for data.*

A page has no real border or visible boundary and has a gray background. The cards have a white background and a visible border and can be seen as different sections within the page. The cards can have different width and height values depending on the size of its content. A first version of the GUI showing the described page and card design can be seen in Figure 5.2. Using this concept allows to have a well defined flow where the user has to go through the shown information from left to right, from page to page and then within a page from top to bottom through the different sections called cards.

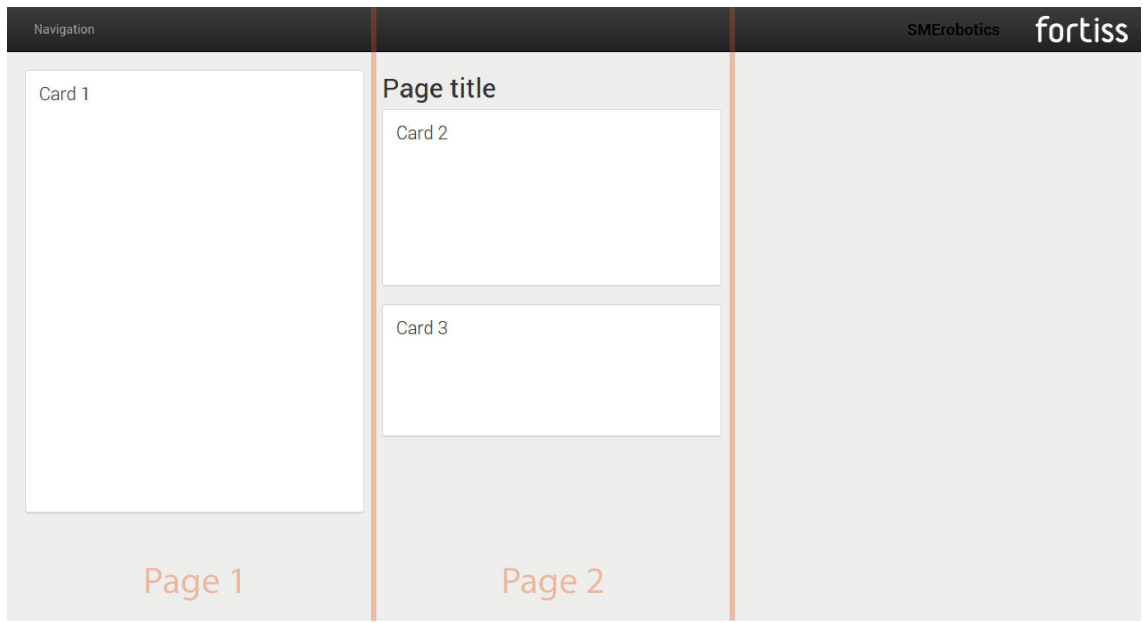

**Figure 5.2:** *Basic design of the GUI: it consists of a variable number of pages. Each page contains any number of cards. Cards are like sections within the page containing different information.* Card 1 *belongs to the first page.* Card 2 *and* Card 3 *are sections within the second page.*

## 5.3  Implementation

When implementing the GUI the main focus lay on easy extensibility for new functions and easy understandable code to provide new developers a code basis that doesn't require much familiarization.

On the server side node.js with express was used. Since the GUI was developed as a Single-Page-Application, there's not much code required for delivering the page contents. It delivers the static files, like .js and .css, and renders the EJS[11] HTML templates for the client.

The client side needed much more implementation effort because most of the logic is implemented within the browser, the server only handles page delivery and database connections. The first step on the client side was to set up a simple backbone.js basis which can be reused for all the different classes. Backbone.js is a Model-View-Controller framework. The Models are JavaScript objects which contain the data displayed by Views. If any attribute changes within this Model, backbone.js automatically takes care of updating the view and re-renders it.

The base class for the application is the `PageApp` class which contains code to manage (add, remove, resize, show, hide) pages in the current view. `ProcessApp` extends this class with additional functions for database initialization.

---

[11]http://embeddedjs.com/

The `PageView` class extends the *Backbone.View* and uses the `PageModel` class to store and retrieve data for creating the page. Each page can contain one or multiple cards. Such a card is represented by the `CardView` class using `CardModel` to store data on how the card should be visualized. A Card contains one or more sections. There are different section types available, which derive from the `SectionView` class. For each listed section view, there's a corresponding model class which is used for storing the data shown within the view:

**TextView**  simple block of multiline text.

**MediaView**  displays an image or video.

**InfoView**  is a combination of text an media view. It shows an image on the left side and a description text on the right.

**ListView**  a list of clickable items. An item can show an image including text, or text only

**CategoryView**  List of clickable items which can be filtered by category

After the `ProcessApp` initialized the database connection and retrieved the required data (like list of processes and tasks) it creates the first page containing the list of processes. To create a page the developer simply has to create an instance of the `PageModel` class. This instance then contains a list of `CardModel` instances for each card. Each `CardModel` again contains a list of all the sections (derived classes of `SectionModel`) within this card. Creating and initializing the corresponding view instances is implemented within each view class. This structure allows easy creation of new pages by just entering the metadata, without bothering with the specific visualization and rendering code.

All the views use an event based communication: if the user clicks on a list item, an event is fired by the `ListView` to the model used for creation of this list. The programmer can subscribe to such an event and then react on it by e.g. opening the corresponding subpage. Figure 5.3 shows the final version of the graphical user interface for programming the robot.

For some of the parameters required by a task, a specific input modality had to be implemented. There's for example a 3D visualization of object models using three.js where the user can select points or edges on the object or define constraints between two objects.

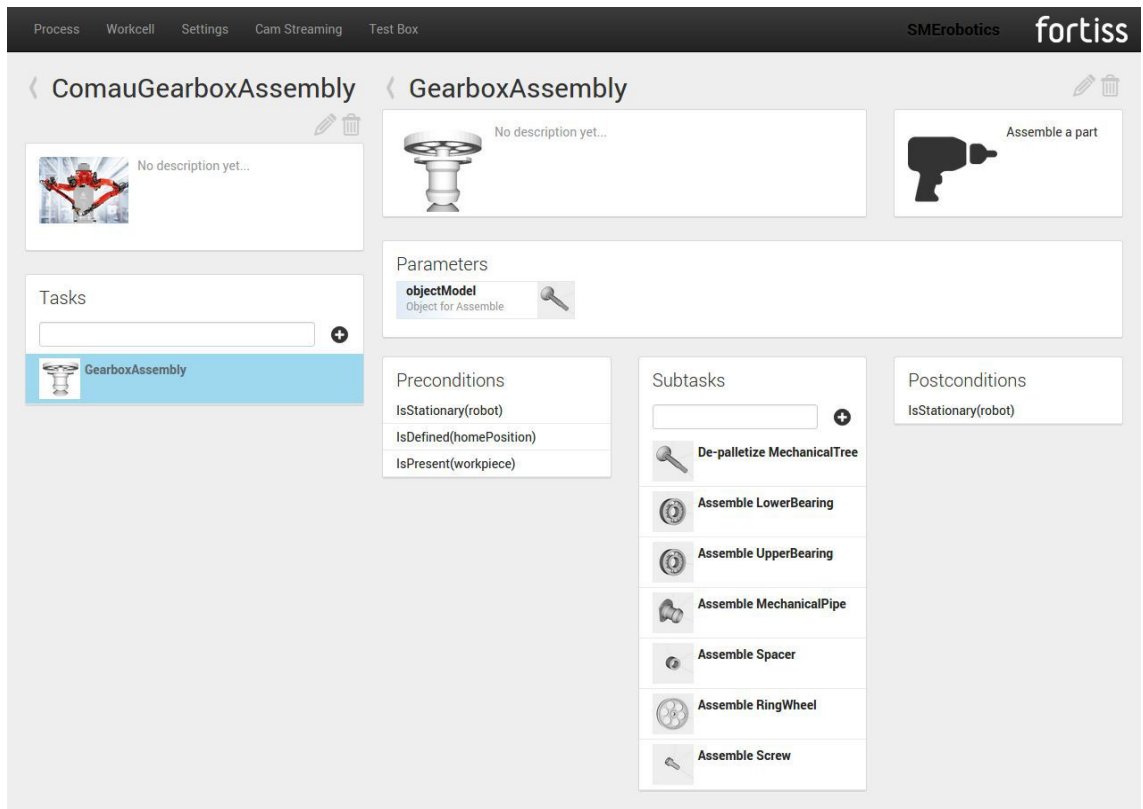

**Figure 5.3:** *The final version of the GUI. It currently shows two pages: on the left is the "ComauGearboxAssembly" task with one subtask. The detail page for this subtask is opened on the right hand side. It shows the parameters required for this task, Pre- and Postconditions and Subtasks of this task.*

## 5.4 Integration

The goal of developing the GUI was to integrate it into the already existing system, developed during the SMErobotics project at fortiss. There are different components, like a semantic storage, a knowledge database, 3D tracking components and robot simulation. All these components use ROS[12] as an interface and communication channel. Therefore the GUI should also integrate ROS so it can communicate with these components and exchange or query data.

Since JavaScript doesn't provide native ROS support, the *rosbridge* node [13] was used. It provides a simple JSON API for non-ROS applications to communicate with other ROS nodes. In combination with *roslibjs* [14], a JavaScript library for connecting to the JSON API, it's possible to subscribe to any ROS Topic, publish data or connect to a service using an event based notification system. As soon as new data is published to a topic, a JavaScript callback function is called with this specific data.

---

[12]http://www.ros.org/
[13]http://wiki.ros.org/rosbridge_suite
[14]http://wiki.ros.org/roslibjs

The GUI uses roslibjs for storing all the data into the database, to fetch data from the semantic storage, and to integrate the different input modalities through it. When the user selects the *Point by hand* input modality, a call to a ROS service is performed to activate the Kinect or LeapMotion sensor. The corresponding software component then publishes continuously, which object the user selected. The GUI subscribes to this topic and then displays this information to give feedback to the user.

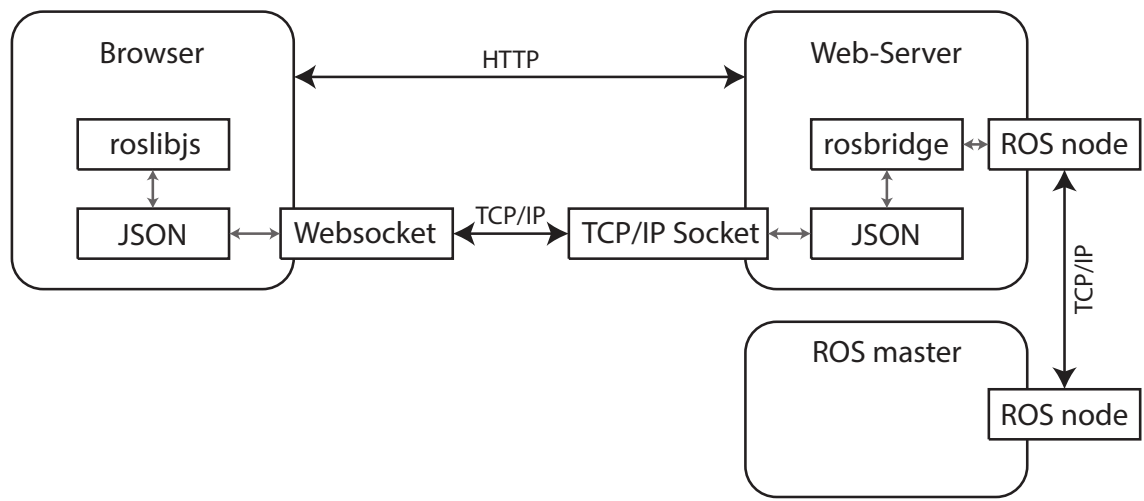

**Figure 5.4:** *Dataflow between the Browser and the Web-Server. The Browser requests the webpage from the webserver using HTTP, then initializes roslibjs and then can call different JavaScript functions from this library. When a function is called, roslibjs generates a JSON data package and sends it through a Websocket to the webserver, where rosbridge opened a TCP/IP socket and listens for connections. Rosbridge also creates a standard ROS node to communicate with other nodes and the ROS master. Published data on subscribed topics is automatically transformed by rosbridge into a JSON package, sent to the client and there an event is called having the data as payload.*

# 6 User Study

The user study should evaluate the usability of multimodal input for task based industrial robot programming, and evaluate the intuitiveness of the different input modalities used in our system setup. It mainly focuses on which input modality the user prefers for different tasks. It uses the graphical user interface presented in previous chapter as the main interaction method with the system.

[Bevan, 1995] defines usability as "quality of use" which has three main aspects: effectiveness, efficiency and satisfaction of users carrying out specified tasks in specified environments. According to [Frøkjær, Hertzum, and Hornbæk, 2000] these three aspects are defined as:

**Effectiveness** is the accuracy and completeness with which users achieve certain goals. Indicators of effectiveness include quality of solution and error rates.

**Efficiency** is the relation between the accuracy and completeness with which users achieve certain goals and the resources expended in achieving them. Indicators of efficiency include task completion time and learning time.

**Satisfaction** is the users' comfort with, and positive attitudes towards the use of the system. Users' satisfaction can be measured by attitude rating scales such as SUMI [Kirakowski and Corbett, 1993].

[Frøkjær et al., 2000] shows that it's important to account for all three aspects because a subset is often insufficient as an indicator of overall usability. Since this user study doesn't aim to evaluate the usability of the system but only the different input modalities and the users opinion on these modalities, the user study mainly focusses on the satisfaction aspect and intuitiveness using the input modalities.

Intuitiveness is defined as "the subconscious application of prior knowledge that leads to effective interaction" [Naumann, 2009]. This definition of intuitive use is captured in the user study by different subscales like *Perceived Cognitive Load* and evaluation of preferred input modalities.

## 6.1 Hypothesis

Before designing or conducting a user study you should have a hypothesis or statement to confirm or contradict by the results of the study. This master thesis evaluates different input modalities for various tasks used in task-based industrial robot programming. For most of the tasks four input modalities are available: gesture input (point by hand), speech input, pen input (using a 3D device) and a graphical user interface for touch input.

The goal of the user study is to find out which input modality is preferred by the user for which type of parameter. The user study is using the Wizard of Oz approach for implementation-independent evaluation. This means that the user thinks the system is recognizing his input (gestures, speech) but instead a human interprets the inputs and forwards them to the system, invisible and unknowingly to the user (see [Dahlbäck et al., 1993]).

Each participant has to try out all four input modalities for the same tasks and afterwards order them based on his preferred use. My hypothesis is that most of the users prefer pointing by hand, followed by pen input, touch input and at last speech input, since for example pointing with the finger to select an object is the most intuitive way, whereas speech input is intuitive, but most of the users won't feel comfortable or don't know how to describe the object to the system.

## 6.2 User Study Design

Each participant had to complete four different phases, which are based on the SUXES evaluation method described in [Turunen and Hakulinen, 2009]. It divides the user study into four phases (see Figure 6.1):

**Phase 1** Introduces the participant to the user study by explaining what the user study is about and gathers some Background Information from the user.

**Phase 2** In this phase the user gets a short introduction to the system by presenting to him the main features. Afterwards he has to fill out an expectation questionnaire.

**Phase 3** The user has to conduct experiments with the system and evaluate his experience afterwards.

**Phase 4** In the last phase the participant has to fill out an opinion questionnaire about different aspects of the system.

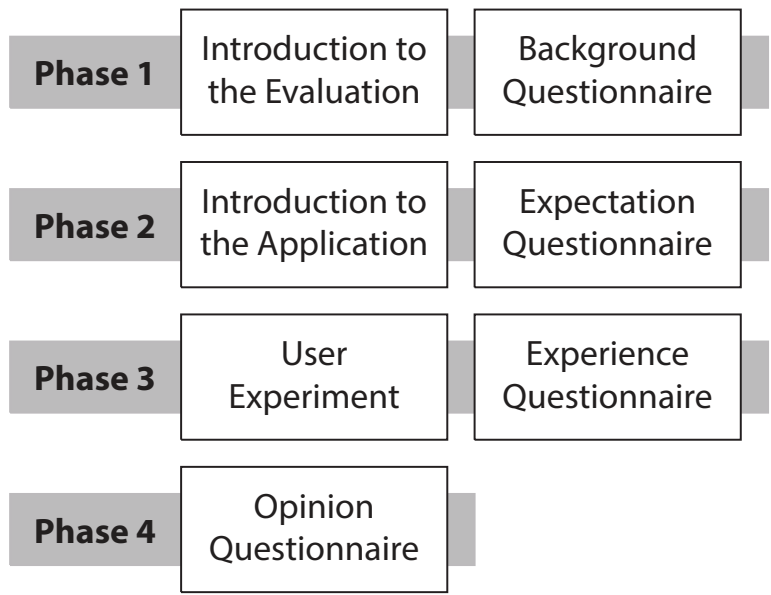

**Figure 6.1:** *The user study is divided into four phases based on the SUXES evaluation method. See also: [Turunen and Hakulinen, 2009]*

### 6.2.1 Phase 1: Introduction to the Evaluation & Background Questionnaire

The first phase of the user study gives a short introduction to the whole user study, how long it will take and what the main goal or hypothesis is. The participant also has to sign an information consent where he agrees, that the collected data will be used for evaluation and that there's a video camera recording the practical parts during the study.

The second part of phase 1 asks the user about some personal information like age, gender, and important for the later evaluation, the expertise in using computers and knowledge in the field of robotics in general (see Appendix A.1.1).

### 6.2.2 Phase 2: Introduction to the Application & Expectation Questionnaire

After filling out the Background Questionnaire, the participant will get an introduction to the system set-up. This should explain all the different sensors, their location and briefly how they process and detect the different input modalities. Since the user study uses the Wizard of Oz approach the user should think that he is really interacting with the system, therefore it's important to have different sensors built into the system, but it doesn't matter if they are working or not.

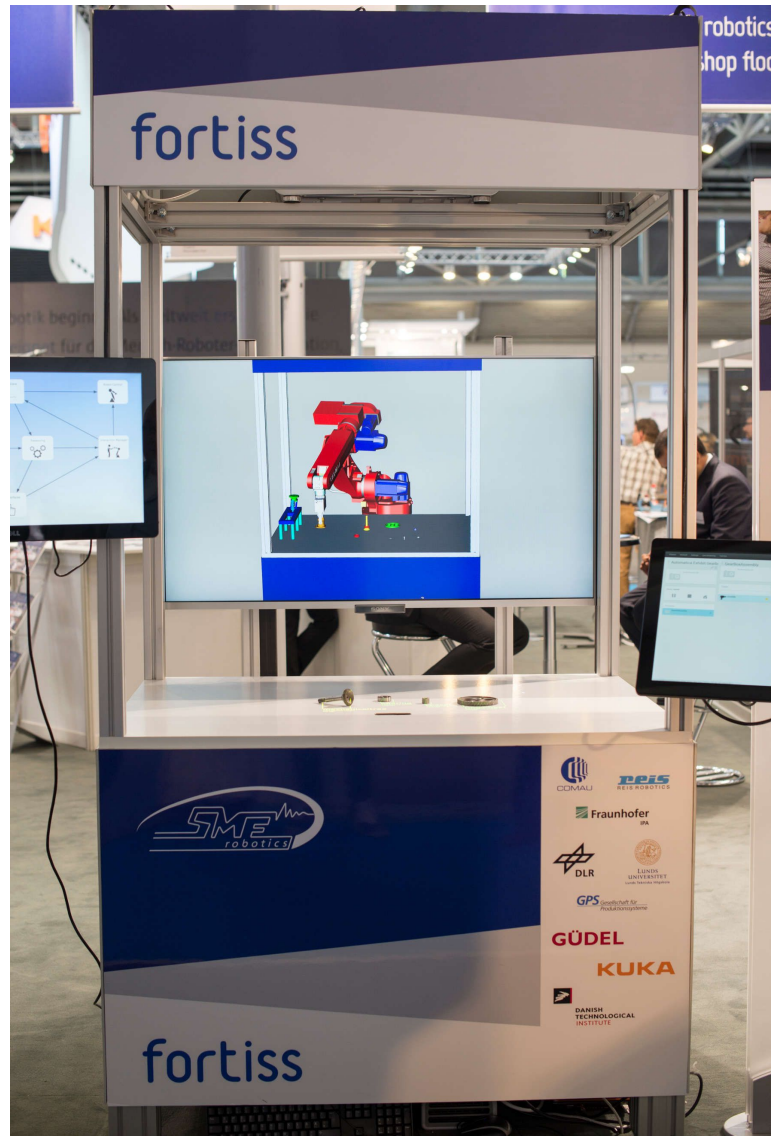

**Figure 6.2:** *This image shows the robot cell or work place where the human coworker programs the robot. In the center is a big screen showing a simulation of the robot. The new version of this cell already uses a real robot instead of the screen. At the right side the touchscreen is mounted, with which the user can program new tasks.*

Our robot cell has different components (see Figure 6.2): the table is used for putting the work-pieces somewhere visible to the user, there's a projector on top of the table hidden within the cage to project object information onto the table, and a touch screen is mounted aside to visualize the current task to program and to give the user feedback on all the input modalities. There's also a (non plugged in) Kinect 2 device mounted on top of the table to pretend the gesture recognition is working. During the introduction it's also explained that the microphone for speech recognition is built into the Kinect sensor. Two infrared cameras for tracking the pen input device are also mounted near the Kinect 2 sensor. These cameras are not used during the experiment since it uses a Wizard of Oz approach.

The touch screen is showing the GUI presented in Chapter 5. To introduce the user to the system, there are three simple tasks which he has to go through. The first task (see Figure 6.3) only needs one parameter of the type object model.

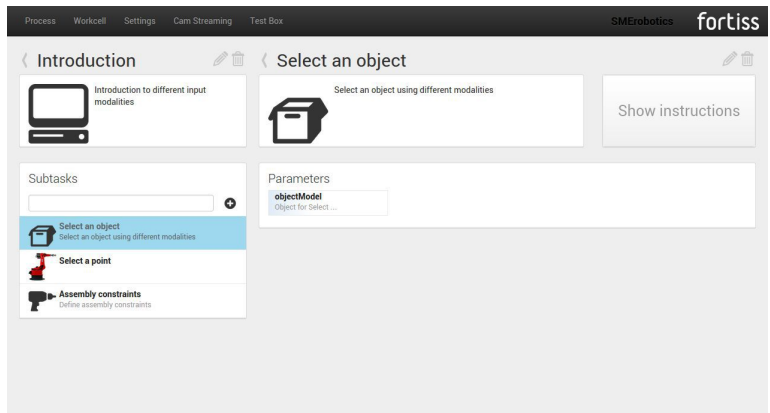

**Figure 6.3:** *The GUI used during the user study. It visualizes the task being programmed and the parameters which are required for this specific task. On the left side there's a list of three tasks used for introduction, where the user has to try out all the modalites. On the right side there's a list of parameters for the selected task. In this case the task requires only one parameter which is of the type* objectModel.

Selecting the parameter, a new window is opened where the user has to select an input modality for how he wants to set the parameter (see Figure 6.4)

The user has four input modalities to choose from:

**Touchscreen** The touchscreen has different visualization types: for object selection it shows a list with all the objects (see Figure 6.5(a)) where the user can select one. Point and edge selection can be done through a touch-friendly 3D visualization.

**Point by hand** When selecting point by hand the user doesn't need any additional device or tool to set a parameter. He simply points with his forefinger to an object an hovers there for about a second to select the object or to select a point on an object (see Figure 6.5(b)). This would require a working Kinect or Leap Motion sensor and the algorithm to track and detect hand gestures. Since the Experiment for this thesis uses the Wizard of Oz approach, there's someone standing behind the participant with a tablet in his hands to manually select the correct value for the parameter based on the participant's input.

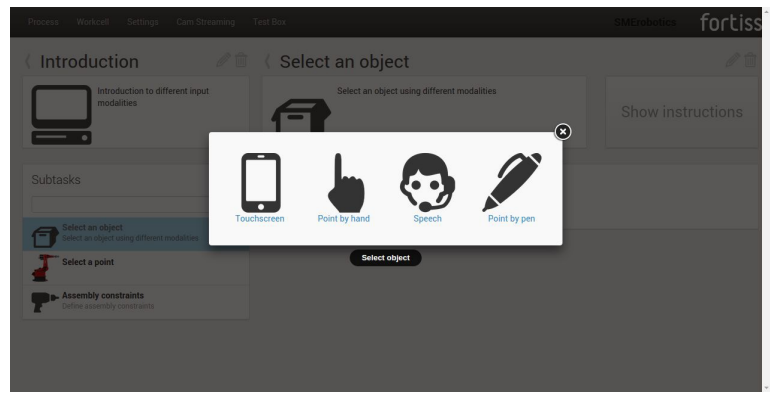

**Figure 6.4:** *By selecting one of the four provided input modalities, the user can define the value for the requested parameter. Possible input modalities to chose are* Touchscreen, Point by hand, Speech *and* Point by Pen.

**Speech** For speech input we explained to the participant that he doesn't need any special vocabulary, instead he should speak to the robot like he would speed to a human worker. The only thing the user has to do, is to begin his sentence or command with "Robot: . . .". These natural commands are possible since we use Wizard of Oz where a human supervisor interprets the spoken commands unknowingly to the participant.

**Point by Pen** For pen input we used the system developed in [Gaschler et al., 2014]. It's a pen-shaped device with infrared tracking balls attached to determine the position and orientation in space (see Figure 6.5(c)). The user has to point to an object or specific location and then press a button on the device to select it. This tracking system is already working but is not yet accurate enough to not influence the results of the study too much. Therefore we used here the Wizard of Oz approach to provide hardware and software independent detection quality.

The remaining two tasks for the introduction are selecting a point on the object for point welding and the last task is to assemble a bearing with an axis by selecting the two objects and then setting the assembly pose as the third parameter for the task. To define the assembly pose, there are two input modalities available to the user: a 3D editor for touch input, and speech input to define the assembly constraints (e.g. "Robot: Put bearing on axis concentric").

After trying all four input modalities and finishing the three introduction tasks, the user has to fill out the Expectation Questionnaire (see Appendix A.1.2). The first expectation question asks the participant on how long he thinks it would take to program a simple pick and place task for industrial robots using systems currently employed in production. The next questions ask the user to order the input modalities according to which input modality he would prefer the most for the following parameter types: select an

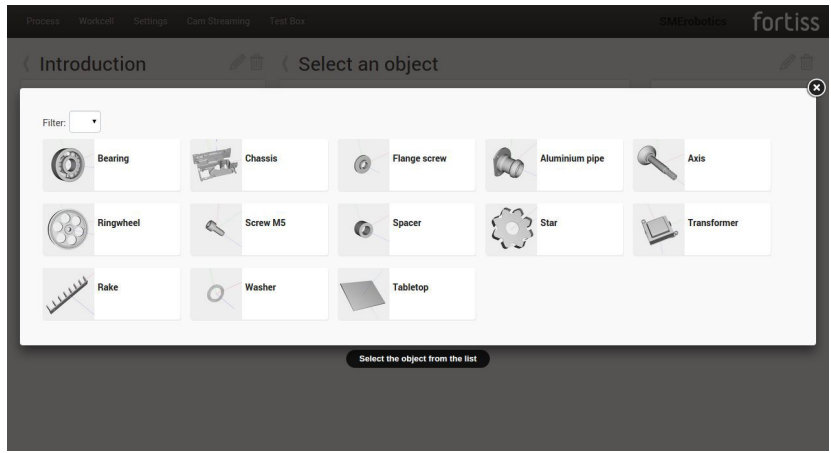

(a) Touch input modality for object model

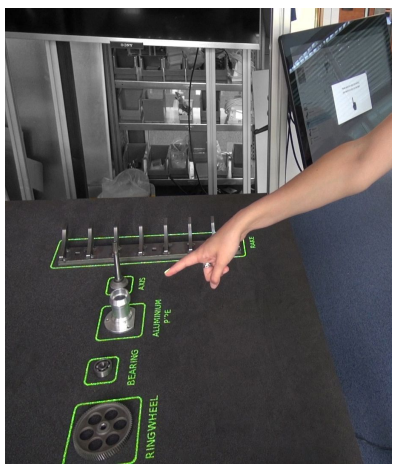

(b) The user points to an object to se-
lect it

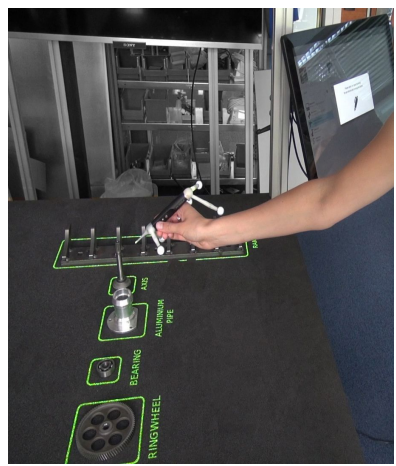

(c) Using the 3D tracked pen the user
can select objects

**Figure 6.5:** *Different input modalities available for the user to select*

object, set a location where to place the object, set assembly constraints, select a point
on the object, select an edge on the object.

## 6.2.3 Phase 3: User Experiment & Experience Questionnaire

After filling out the Expectation Questionnaire from Phase 2, the user has to program
multiple tasks using all four input modalities sequentially. First the participant has to
program all tasks with touch input only, then the same tasks with hand input, then
speech and at last pen input. The four tasks are:

**Pick & Place**  Select an object an place it on the center of the table. Required parameters: object to pick, location to place

**Assemble two objects**  Select the bearing and put it on the axis. The bearing should be placed concentrically onto the axis. Required parameters: object to pick, object to place on, assembly constraints/pose

**Weld point**  Select the rake object and weld a specific point (third from bottom left). Required parameters: object to weld, point on object

**Weld seam**  Select the rake object and weld a specific seam (third from bottom). Required parameters: object to weld, seam on object

Programming these four tasks using all input modalities sequentially, takes in total about 15 minutes. After this practical part, the participant has to fill out the Experience Questionnaire (see Appendix A.1.3). The first question asks the user to order the input modalities based on the experienced cognitive load. This means that the user has to indicate for which input modality he had to think the most. The next 5 questions are exactly the same as in the Expectation Questionnaire, to compare the expectation with the experience after he conducted the experiment.

## 6.2.4 Phase 4: Opinion Questionnaire

The last phase of the experiment is an Opinion Questionnaire (see Appendix A.1.4) where the user has to express his opinion on the different input modalities and the whole system. This includes questions like "It's intuitive to use touch input" or "It was difficult to understand how to use the different input modalities". The goal of this questionnaire is, to understand how the user feels about the system and what he thinks on the input modalities.
The second part of the Opinion Questionnaire asked the user to imagine that he's working in a factory and his job is to program industrial robots. Now he should express his opinion on using the system for exactly this job, by answering different questions like "Using the system would improve my job performance" or "Using the system would make it easier to do my job". The evaluation of these questions should give feedback on the usage of multimodal task-based industrial robot programming for production usage and for Small and Medium-sized Enterprises.

## 6.2.5 Open discussion

The experiment concludes with an open discussion where each participant can give additional feedback on the questionnaires or the system, and can suggest improvements. This part also includes the reveal that the whole experiment is a Wizard of Oz experiment, and that the components of the system weren't actually working, but the instructor of the user study controlled and set the input parameters. This open discussion is a valuable part of each user study because you get worthy feedback and other ideas of which you didn't think at first.

# 7 Evaluation

This chapter evaluates the questionnaires of the conducted user study described in previous chapter. It shows that the hypotheses previously defined is confirmed and that most of the participants preferred hand input whereas speech input is the least preferred one. The evaluation of the opinion questionnaire gives an impression on how the user felt during the user study and what he thinks about multimodal input technologies.

The results of this user study are also integrated into the ontology described in Section 4.5 to extend the knowledge of the system. It's able to select the most preferred input modality for each parameter based on the results of the user study and therefore providing the best user experience possible.

The answers of all the participants for all the questionnaires are also available for download as an *.xlsx* file on the following website: http://profanter.me/publications.

## 7.1 Participants

The first questionnaire asked the participant about background information like age, gender, expertise in using computers and knowledge in robotics. There were 31 participants at the user study, whereat the first participant was used as a tester for the whole setup and therefore the data from this participant isn't included in the data. Consequently the data includes answers from 30 participants. This number of participants should provide a confidence interval reasonably tight as stated in [Nielsen, 2006]. For each user it took about 30 minutes to complete the questionnaires, including the practical part.

Most of the participants were students of different technical and non technical fields and researchers in the field of robotics and embedded systems. Therefore the data of this user study is based not especially on domain experts (i.e. employees in Small and Medium-sized Enterprises) but on a widespread basis of knowledge of different people.

The average age over all participants is 27 years with a standard deviation of about 5 years (max: 41, min: 20). Figure 7.1(a) shows that 23 participants (77%) were male and 7 participants (23%) female. The questionnaires were available in two languages, German and English, to make it easier for the participants to answer the questions, and to make sure the questions are understood correctly. 23 users selected the German questionnaire and 7 users the one in English.

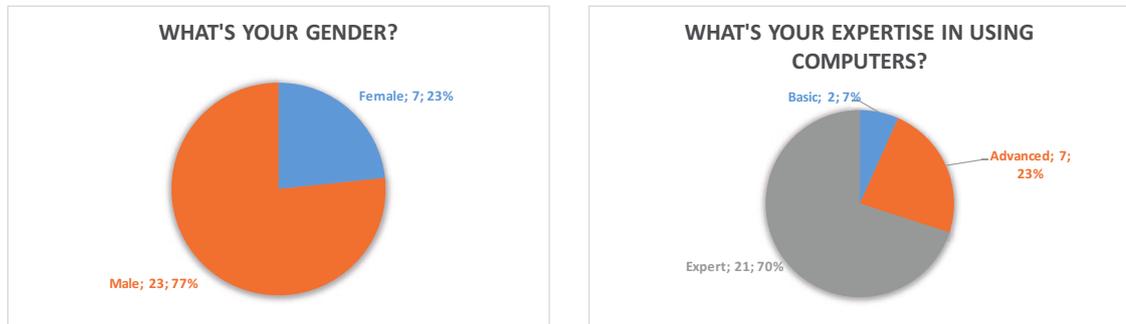

(a) Gender distribution of all the participants of the user study. 23 were male and 7 female.

(b) 2 participants indicated that they have a basic knowledge on using computers, 7 advanced, and the majority were computer experts (either computer scientists or programmers).

**Figure 7.1:** *Background information on the 30 people who participated in the user study.*

To get a better understanding on the knowledge of each participant in robotics and industrial robot programming, there were two questions to answer (see also Questionnaire A.1.1): How much do you know about robotics? And did you already use a TeachPad to program a robot?

The answer distribution for both questions is shown in Figure 7.2. It can be seen that 50% know a lot about robotics (studied robotics, computer science or similar) and about 40% know what a TeachPad is (see Figure 1.3 on Page 7) or used it already to program a robot.

## 7.2 Expectation

After finishing the practical part of the user study, each participant had to answer some questions, if using the input modalities and the system met his expectation or not (see Appendix A.1.4, question 18). Figure 7.3 shows the average value and standard deviation for each answer. As it can be seen the system outperformed the user's expectation for most of the components in terms of simplicity. Only touch input behaved as expected since it wasn't different than using a smart-phone or any other touch device. Touch was also the only component where Wizard of Oz wasn't used and thus represented the

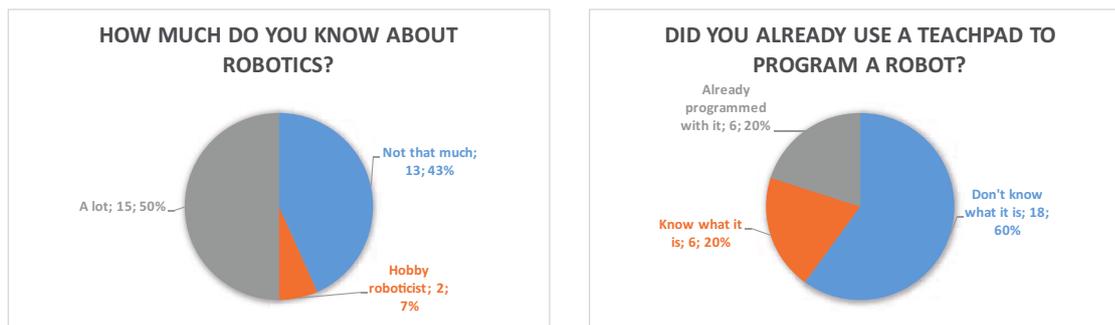

(a) 13 participants answered to the question on how much they know about robotics with "Not that much (heard or read about it)". 2 indicated that they are hobby roboticists and 15 (50%) said that they know a lot about robots (studied robotics, computer science or similar).

(b) 60% of the participants didn't know what a TeachPad is or how it works. 20% indicated that they know what it is and how it works, but never used one. Another 20% already programmed an industrial robot with a TeachPad.

**Figure 7.2:** *Knowledge of the participants in the field of robotics and using a TeachPad.*

current state of the art. The interpretation of all the other components (speech, gesture, pen input) used the Wizard of Oz approach where a human set the values for each parameter based on the participants input, unknowingly to the participant. This may also explain, why for most of the users, handling those input modalities was easier than expected since it's not yet possible to implement speech recognition or gesture recognition which reaches the same detection accuracy as a human.

Another interpretation of this graph is that most of the participants know or think that programming an industrial robot is very complex and complicated, and by using this system they were surprised how easy it can be, if the robot or the intelligence behind the robot can infer information automatically and adapt to many different setups.

Looking at the data separated by gender, there's a significant difference for speech input: female participants selected an average value of 1.7 compared to male participants who have an average of 2.3. This means that for female participants speech input was significantly easier than expected compared to male participants. This also somehow confirms the statement from [Wasinger and Krüger, 2006] that women feel less secure and confident using speech in public environments.

Question 6 of the expectation questionnaire (see Appendix A.1.2) asked the participant to think about a simple pick and place task, where the robot has to pick up an object from a palette and place it on the table with a specific orientation. Then the participant should estimate, how long it would take to program this task to a robot using current systems employed in production. The average over all 30 estimates is about 78 minutes with a standard deviation of about 128 minutes (maximum value: 600 minutes, minimum: 1 minute). The high variation between the values may be caused by the inaccurately stated question, since it is not totally clear if the programmer is allowed to use predefined macros (which would speed up the teaching significantly) or if he has to start from scratch.

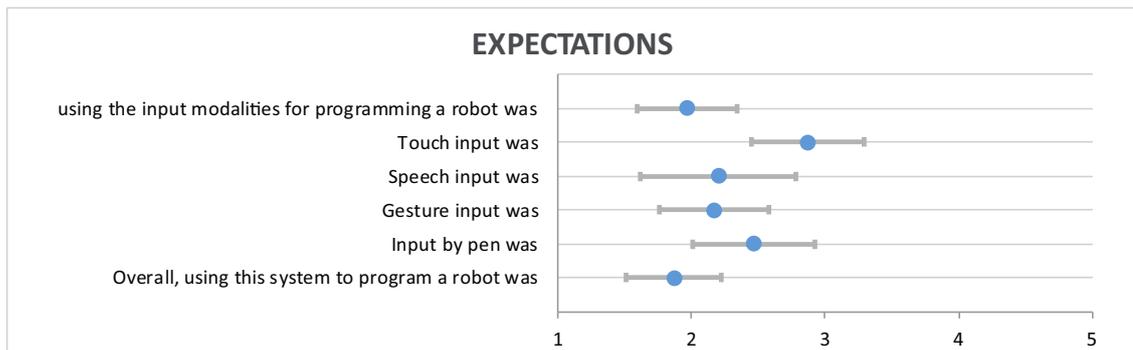

**Figure 7.3:** *This diagram shows the answers for each listed question if it matched the user's expectation. 1 = a lot easier than expected, 3 = as expected, 5 = much more complicated than I thought*

Nonetheless an average of 78 minutes is quite a big estimate. Asking an expert who daily programs and teaches tasks to industrial robots stated that he would need about 15 minutes (using predefined macros) to program such a task.

## 7.3 Input modalities

After a short introduction to the system and quickly trying out the different input modalities, the participant had to order the four input modalities *touch input*, *speech*, *pen input* and *point by hand* according to his preference. After this step, the user had to program four different tasks using all four input modalities sequentially, and then in the next questionnaire, order the input modalities again according to his previous experience. This is described in more detail in Section 6.2.

Figure 7.5 shows the results of the ordered input modalities which will be analyzed here in more detail.

The first question was about the perceived cognitive load for each input modality. The participant had to order the four modalities according to where he had to think the most to use the modality. Figure 7.5(a) shows that *speech input* requires the most thinking whereas *point by hand* was selected by most of the users as the easiest one. The main reason for this result may be due to the fact that most of the participants didn't really know how to talk to a machine or a computer. Even if the system was built to understand everything (Wizard of Oz Experiment), the users were in doubt about which vocabulary they can use, despite they were told that they can talk to the machine like they are talking to a human. This is also related to the findings by [Dahlbäck et al., 1993], who found out that dialogs between human and human differ significantly from human-machine dialogs. He also describes different factors that influence how we speak to a machine or human.

Comparing the cognitive load between the seven female and 23 male participants, there's a significant difference between touch input and input by hand: female participants have an average value of 3.2 for touch input and 2.7 for hand input, which results in in the following order of cognitive load: speech input, pen input, input by hand and the less cognitive load touch input. This means that for female participants using touch input was easier than input by hand. In contrary for the male participants input by hand has an average value of 3.4 and touch input 2.4 resulting in the order speech, touch, pen and then hand input, thus input by hand required less thinking than touch input for most of the male users.

Speech is a very diverse and versatile input modality where you can describe something in a lot of different ways. Hand input on the other hand has more or less only one possibility to tell the robot which object to pick: by pointing to this specific object. This can be a reason why most of the users said that they had to think less when using hand input. Pen input and Touch input are more or less on a par in the middle range.

For the next five questions the user had to order four input modalities based on his experience in the previous practical part of the experiment. A value of 1 means that the user preferred this input modality and put it therefore at the first place. Figure 7.5(b) to 7.5(f) shows the answers for each parameter type which will be analyzed here in more detail.

The *Select an object* parameter required the user to set an object model using one of the four available input modalities. By most of the users *Point by hand* was selected as the most or second preferred input modality (Figure 7.5(b)). The other three modalities are in average between 2.5 and 3 whereat *Speech* has the highest standard deviation of about 1.2 points. Looking at the raw data of the study, Speech was often set to the fourth place but also a few times at the first place resulting in this high standard deviation. The main reason for this result may be, that for Point by hand the user doesn't has to think a lot how to use the input modality. Compared to touch input he has to search for the object within the list which slows down the interaction, for speech input the user has to know the name of the object, and pen input requires an additional yet unknown tool to use and handle. For speech input there's a noticeable difference between the expectation and experience questionnaire: before the practical part users thought that they don't like speech input, but after using it, speech gained a quite better average value than before. This may be due to the fact that most people have a prior knowledge or the impression that speech input doesn't work good enough yet, but after finishing the experiment they saw that speech recognition is better than they thought (because it was a Wizard of Oz experiment where a human interpreted the speech commands, which the user didn't know).

Looking at the *Select an object* parameter results separated by gender, there's a significant difference between male and female participants for the touch input and speech input (see Figure 7.4). For female participants touch input was much more preferred (av-erage value of 2.1) compared to the male participants (average of 2.7) in favor of speech input which was put by most of the female participants on the last place.

Separating the data based on the answer for the expertise in using computers (see Figure 7.1(b)) into two groups, experts and non experts (includes basic and advanced), shows that experts gave *Point by hand* a higher rating (average 1.4) compared to non experts (average 1.9). The reason for this may be that experts already used touch and speech input a lot more times, and know the limitations of those input modalities. Therefore they preferred point by hand over all the others since it is much more intuitive and has a high potential to be improved.

Evaluating the data separately for users that already used a TeachPad shows also a significant difference for the *Select an object* parameter regarding the expected preferred input modality (answers before the practical part) and the finally preferred input modality. The TeachPad participants expected that they won't like speech input (average value of 3.4). After the practical part the average position for speech input as preferred input modality moved from 3.4 to 2.4 which moved speech input from least preferred to the second most preferred input modality. This is most probably caused by the fact that those users thought that speech input is not really suited for programming a robot in any way but then after using it to set the object parameter, they got more convinced and, given that speech input recognizes the commands perfectly, it may be a helpful addition.

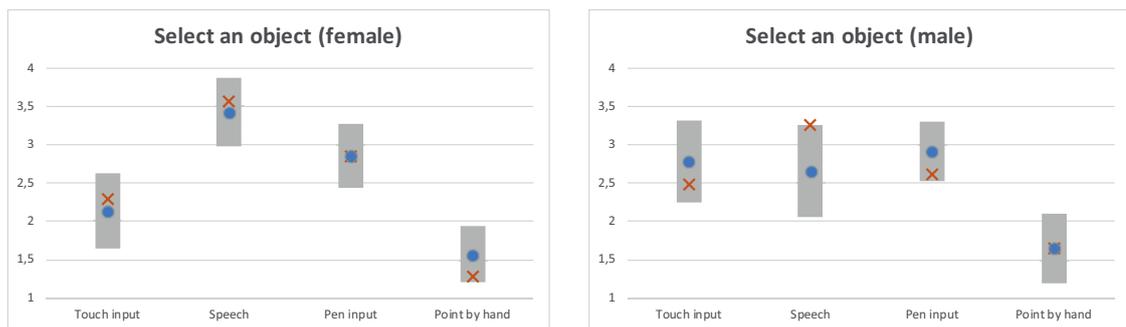

(a) Female participants gave touch input a significantly higher rating.

(b) Compared to female users speech input is more preferred by male users.

**Figure 7.4:** *Preferred input modalities for* Select an object *parameter separated by gender. 1 is the most and 4 the least preferred modality. The blue circle indicates the average over all participants in this group, the gray bar shows the standard deviation for each modality. The orange cross is the average from the expectation questionnaire where the user had to order the modalities according to which he will most likely prefer the most.*

For the *Set location* parameter (Figure 7.5(c)) the choice of the participants is quite different. At the first place is *Touch input*, followed closely by *Point by hand.* Pen input has approximately the same average value as in the previous parameter, speech input is even worse. During the open discussion at the end of the study some mentioned that for them an influential factor for choosing a specific order of the modalities was the time you need to switch between touch and other input modalities. This may have played a bigger role for the location parameter since the visualization of the table on the touchscreen was simply a 2D rectangle where the user had to touch a specific location. Staying on the touchscreen is much faster than turning to the table and pointing to a specific location. *Speech* is at the last place since it is more complicated to describe a position accurately compared to simply pointing at it. *Pen input* is approximately in the same range as for the previous parameter, one reason for that may be that the user has to pick up the pen first, before pointing to a location, which is slower than just pointing with a finger.

To define *assembly constraints* only two input modalities were available: *Touch input* and *Speech*, since defining constraints using gesture or pen input is not that intuitive or not even possible. For Touch input the user was presented with a simple 3D editor where he could move parts around and define relations between parts like concentric, coplanar or a specific distance. For example selecting two cylinders by touching them and then pressing the concentric button, defined a concentric constraint between those two parts. 25 of the 30 participants preferred Touch input over Speech input. Observing the user study showed that more or less all participants struggled to find an easy description for putting the bearing on the axis and aligning the cylinders concentrically.

*Select a point* and *Select an edge* are two parameters that needed to be set for the welding task. The problem specification was to set the third point/edge from bottom left as the welding position. For touch input the user had a 3D visualization of the object where he had to touch the correct edge or vertex to select it. The results for point and edge are nearly identical (see Figure 7.5(e) and 7.5(f)): pointing by hand is the preferred way for setting this parameter followed by pen input. The main reason why pen input is before touch and speech may be, that it is more intuitive to use a sharpened object to accurately define a position or edge in 3D. An additional curtail of touch input in favor of pen input was the not perfectly working 3D visualization, because it had some minor bugs impairing user inputs. Using speech only, to define a specific position in 3D on an object is quite difficult and the majority of the participants were unsure how to describe it to the robot, one participant event refused to try it, because he said that it doesn't make any sense to define such a position using speech.

Nearly all participants which indicated that they are experts in using computers, didn't really like speech input for setting a point or an edge. 20 out of the 21 experts set speech input to the last place resulting in an average position of 3.9 among the experts.

# 7.4  Opinion

The opinion questionnaire (see Appendix A.1.4) asked the user about intuitiveness of the modalities and general interaction with the system. It includes questions from the After-Scenario Questionnaire (ASQ) [Lewis, 1995] and the Software Usability Measurement Inventory (SUMI) [Kirakowski and Corbett, 1993].

The goal of this questionnaire was to get a basic impression what the user thinks about the system and the different input modalities. Figure 7.6 shows the distribution of the answers from all 30 participants.

The graph shows that on average users had fun using the different input modalities, and they were satisfied about the ease of completing the tasks and the required time it took to complete them.

Two interesting questions within this questionnaire are "The system should not allow to select a modality. It should pre-select the most suited one for each task." and "I feel more secure if there are less modalities to choose". The first question has an average value of 3.5 which means that in average the participants neither agree nor disagree. Looking at the raw data shows that 5 participants selected option 6 (strongly agree), only one participant selected option 1 (strongly disagree) and 9 selected option 2 (disagree). Giving these options a bigger weight it means that approximately 10 participants want to select the input modality on their own whereas only 6 say that only the most suited one should be available. A conclusion of this result may be to reduce the amount of available modalities to only two, so the user can still select the one he prefers but doesn't has to decide himself for one out of four input modalities.

When separating the data by the answer for expertise in using computers, the picture gets even more clear: experts somewhat agree (average of 4.0) that the system should preselect the most suited modality whereas non expert users would like to select the input modality on their own (average value 2.3, disagree to somewhat disagree). The same trend can be seen if the data is separated into two groups: participants who know a lot about robotics and the other group consisting of hobby roboticists and participants who don't know that much about robotics. Experts or daily users of the system likely tend to rush through the programming steps. If they need to select the modality for each step it slows them down, but if they know that for example selecting an object they have to use input by hand, it makes their interaction with the system much more efficient.

Looking at the subquestions regarding the intuitiveness of different input modalities, it can be seen that speech input, and keyboard and mouse are seen as less intuitive by most of the users. Point by hand is the most intuitive input modality even before pen input and touch input.

For the second part of the opinion questionnaire the participant had to imagine that he's a factory worker and programming industrial robots is his daily job (see Figure 7.7). Summarizing all the questions and answers, the overall statement is that the participants would find the presented way of programming an industrial robot very useful for their job, and that it would increase their job performance. Looking at the raw data for each participant and each answer, all participants selected at least *somewhat agree* (4) or more (for all questions except number 4, 9 and 11). This means that all participants think the system is an improvement for their job.

Comparing the answers for "Learning to operate the system would be easy for me" between experts and non experts, shows that experts find it easier to learn how to operate the system (average value of 5.3, agree to strongly agree) compared to non experts users (average of 4.9, agree).

It would be interesting to ask some workers whose job really is programming industrial robots, and check if they are also convinced about the system. Unfortunately this wasn't possible for this thesis but such an experiment is already planned for next year as part of the SMErobotics [1] project.

For participants who already used a TeachPad, or at least know what it is (see Figure 7.2(b)), an additional question was shown, which asked the user to estimate, how much time saving (in percentage) can be achieved by using the input modalities presented for this system, compared to using a TeachPad. The average value over all 12 answers is about 71% (standard deviation: 23%) which means that they think using multimodal input would speed up the whole teach in process by 71%.

## 7.5  Video analysis & Open discussion

During the practical part of the user study, a video camera recorded the participant's interaction with the system to analyze it afterwards. After the study there was also an open discussion with each user to get additional feedback aside from the questionnaires. This section summarizes observations made when watching the recorded videos and evaluates comments from the open discussion.

A main observation which could be seen for almost all participants was that they had problems using speech input: the users didn't know how to explain something to the robot including all the required details. Since the experiment used the Wizard of Oz approach, it didn't really matter how the user said something, because a human supervisor interpreted the commands and set the correct value for the system, whereas the users of course didn't knew about that. Nonetheless they said that they don't really know how to talk to the robot because they thought that it has a reduced set of vocabulary or doesn't understand everything. This may be due to the fact that most of the participants already used speech input for a device (e.g the Smartphone) or at least heard or read about it and therefore were biased. After telling them that they can use any kind of vocabulary they felt a bit more secure, but still it wasn't that easy as talking to a human counterpart. One participant said that using speech input is odd, but maybe after using it for a longer period it may get easier and faster. Looking through the recording has shown that this is already the case after using speech only one time: during introduction the participant had to define assembly constraints using speech, where most of the users struggled to give the corresponding command but then, the second time they had to define similar assembly constraints, the user was significantly faster.

---

[1]http://www.smerobotics.org

When using speech input, it also makes quite a difference if the user is a domain expert or not: for the assembly task, where the robot hat to put the bearing on the axis and align it concentrically, most of the users gave a command similar to "Robot: Pick bearing and put it on axis concentrically". In contrary one participant was a mechanical engineer who gave a more precise description using special vocabulary from his domain of expertise: "Robot: Put bearing on shaft to collar" which exactly defines that the bearing must be pushed on the shaft until it touches the collar.

Another interesting observation was that quite a few participants always looked to the microphone when giving speech commands. During the introduction to the system, the location of the microphone was shown (it was located within the Kinect sensor on top of the table). So instead of just looking at the touchscreen (which was mounted on the left side of the table) and giving speech commands, they turned to the table and looked upwards to the Kinect sensor. When designing such a system the developer should keep this in mind and mount a microphone near the touchscreen or show some animation or feedback on the touchscreen to keep the user looking at the touchscreen and thus saving time.

One participant said that he didn't like hand input that much because the feedback was missing. The system that was used during the study, didn't provide any visual feedback directly on the table, if the object was selected properly. The user always had to look at the touchscreen to check it. This may also have slightly influenced the overall result of the experiment.

Another participant also noted that if you are looking already at the touchscreen and you have to decide, whether you choose touch as input, or another input modality where you have to move away, it is more likely or consequent to select touch over the other modalities.

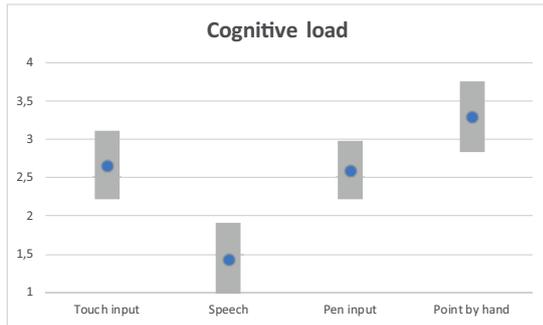

(a) Input modalities ordered by cognitive load (where did the user had to think the most, required the most attention). 1 means high cognitive load, 4 less cognitive load.

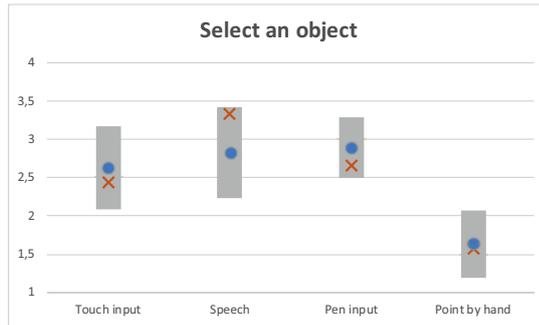

(b) For selecting an object the users preferred Point by hand. It's noticeable that speech input was least preferred in the expectation questionnaire compared to the experience.

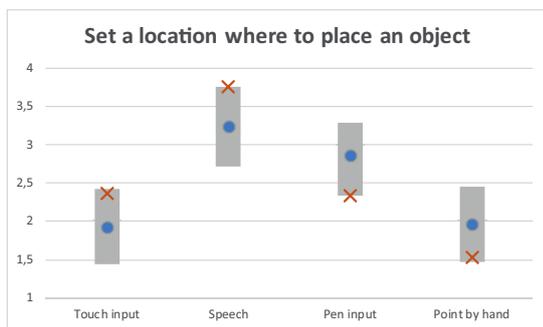

(c) Touch input is the preferred modality for setting a specific location on the table closely followed by gesture input.

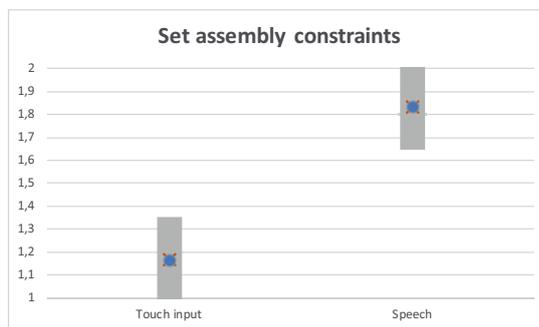

(d) 25 of the 30 participants found Touch input easier than Speech input because it's quite difficult to describe assembly constraints using speech.

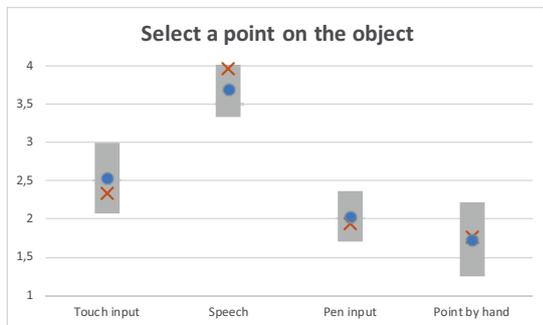

(e) Selecting a point on the object by pointing with a finger was preferred by most of the users. Pen input is right after Point by hand.

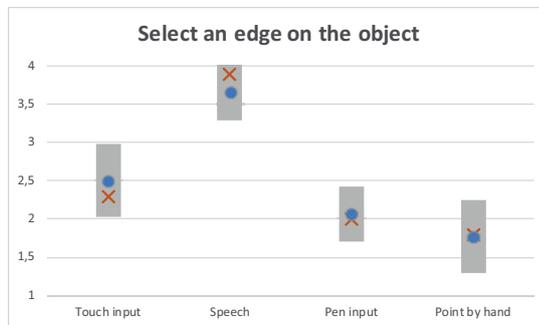

(f) The answers for selecting an edge are nearly identical to selecting a point on the left side.

**Figure 7.5:** *Graphs showing the results for each question where the user had to order the input modalities for the different tasks or cognitive load according his experience in previous practical part. For the cognitive load, 1 means high load, 4 less cognitive load. For all the other graphs 1 is the most and 4 the least preferred modality. The blue circle indicates the average over all 30 participants, the gray bar shows the standard deviation for each modality. The orange cross is the average from the expectation questionnaire where the user had to order the modalities according to which he will most likely prefer the most.*

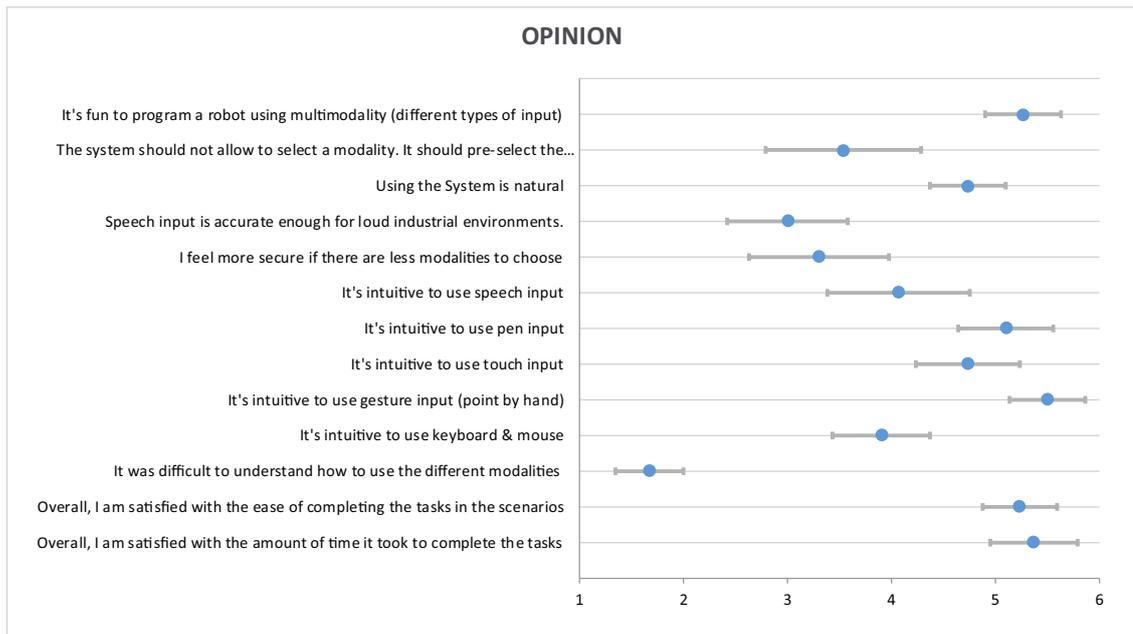

**Figure 7.6:** *Distribution of the answers for questions about the intuitiveness of the input modalities and opinion about the whole system. 1 = strongly disagree, 2 = disagree, 3 = somewhat disagree, 4 = somewhat agree, 5 = agree, 6 = strongly agree. The blue dot marks the arithmetic mean of all 30 participants, the gray line indicates the standard deviation.*

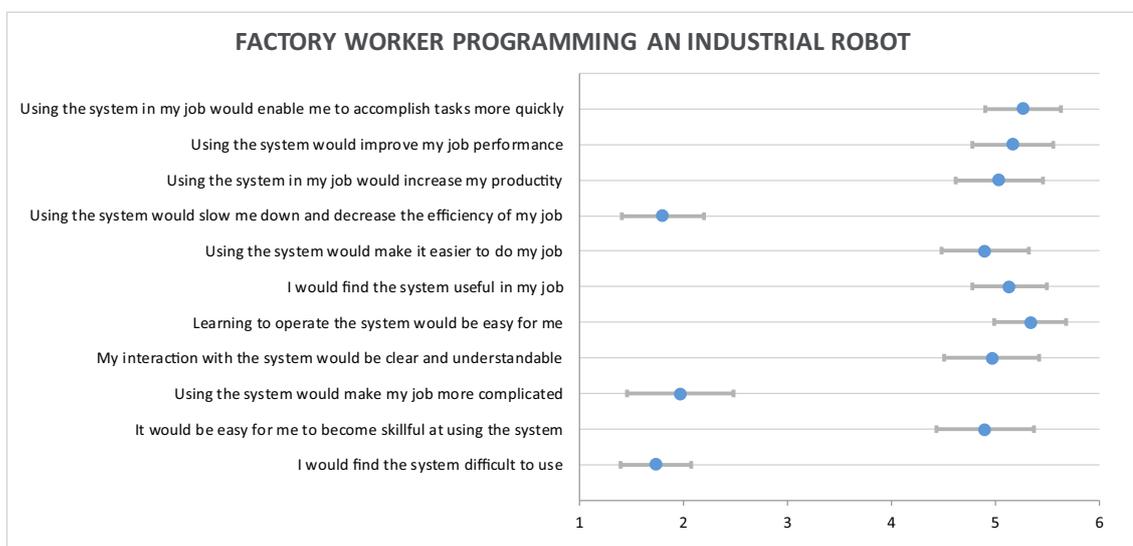

**Figure 7.7:** *To answer the questions above, the participants had to imagine that they are factory workers and programming industrial robots is their daily job. 1 = strongly disagree, 2 = disagree, 3 = somewhat disagree, 4 = somewhat agree, 5 = agree, 6 = strongly agree. The blue dot marks the arithmetic mean of all 30 participants, the gray line indicates the standard deviation.*

# 8 Conclusions

This master thesis focused on the implementation and evaluation of multimodal input and output channels for task-based industrial robot programming.

After listing and evaluating different human input and output channels, computer input and output modalities were explained in more detail, including historical background and possible application areas.

The presented task-based programming is based on the hierarchical concept of process, task and skill. Skills are primitive functions provided by the hardware or actions requiring additional atomic steps or sensor functions. Tasks is an abstract work description that maps to one or multiple skills. A task can have different skill mappings providing the same functionality, thus there may exist multiple skill implementations for the same task that use for example different sensors or algorithms. A process is an abstract description of a robot program consisting of one or multiple tasks. Different task types in four application domains (assembly, welding, woodworking, metal processing and bonding) were analyzed and described in more detail: Which skills are required for the task? What are the parameters the user needs to set? Which input modalities can be used? What are the uncertainties or problems which may arise when executing the task?

Based on this task analysis, an ontology was constructed which semantically describes the relation between the tasks and corresponding skills, which parameters are required, of which data type they are and which input modalities can be used to set those parameters. It also includes representations of hardware and software system components like sensors or specific algorithms. This ontology can be used for cognitive robot cells, for example to automatically infer, if specific input modalities are available based on the equipment of the cell or which modality should be used for a specific parameter type.

For programming the tasks and evaluation within the user study, an intuitive graphical user interface using modern web technologies was developed, which is optimized for touch input and integrates multiple input modalities.

The goal of the user study was to evaluate which input modalities users prefer for which type of parameter. The study was divided into four phases: background information, expectation, experience and opinion. These phases evaluate different aspects within multimodal interaction: what the user expects from the system, how he experienced it and what his opinion is.

After filling out a questionnaire the participants had to program two assembly and two welding tasks to the robot using all four input modalities: speech, gesture input, touch input and pen input sequentially. So each user had to use each modality for each task. The tasks required various types of parameters, which the user had to set: object models, location on the table, assembly constraints and vertices or edges on an object.

Evaluating the results of 30 participants clearly shows that most of the users prefer gesture input over touch input, pen input and speech input. For selecting a vertex or edge, pen input was slightly more preferred compared to other parameter types. This is similar to the results from [Ren et al., 2000] where users mainly preferred pen input for CAD applications. Additionally the study confirms the results from [Wasinger and Krüger, 2006], who stated that women feel less secure and comfortable than men when using speech input.

The study also reveals that speech input on its own is not very suitable for usage in task-based robot programming. Due to the high diversity when using speech, there are multiple possibilities to define the identical command, using different words or sentences. During a Wizard-of-Oz experiment that's no problem, but implementing such a speech recognition system is quite complex, if it is even possible. For example position indications (left, right, front, back) are not defined accurately: is it defined from the user's point of view or from the robot. Minimizing these drawbacks requires special training for the user to sensitize him regarding these problems, which isn't the goal of intuitive and easy use of an interface.

[Naumann, 2009] state that prior knowledge or familiarity with a specific input modality influences the rating or preferred choice of modality. Looking at the results of the user study for this master thesis, this statement couldn't be confirmed: nonetheless most users are quite familiar with touch interfaces, input by pen gained a similar high or even better rating than touch input, even though most of the users didn't yet use this kind of input modality.

Another conclusion drawn from this user study is to reduce the amount of available input modalities for a specific parameter: experts stated that the intuitiveness of the GUI would increase if the system preselects the most suitable input modality whereas non-expert users tend to try out and play with the system, using different input modalities.

The outcome of this thesis clearly motivates further research in multimodal interaction for robot programming as it provides the user with greater naturalness, expressive power and flexibility. The participants state that using multimodality makes robot programming more intuitive, easier and faster.

## 8.1 Future Improvements

The evaluation of the current system showed that there is still space for improvement. For this thesis, fusion of input modalities wasn't used and implemented due to time constraints, but using multiple modalities at the same time will most probably increase the intuitivity of the whole system and make the programming of industrial robots easier. For future user studies this should be considered and should also include some questions about modality fusion, for example if the user would have preferred to use gesture in combination with speech or speech only.

Within the study the user always had to select the input modality on the touch screen to use it. As some of participants encouraged, a huge improvement for the system would be to automatically select the most suited modality based on the current parameter and knowledge of the system about the world state. The ontology created within this thesis is a good starting point for this improvement. Additionally the system should detect a switch between input modalities on the fly, without explicitly selecting one, or telling it to the system.

During the Wizard-of-Oz study the system didn't provide much user feedback: e.g. when using gesture input to select an object there was no visual feedback on the table, but only on the touchscreen. Future improvements should focus on how additional feedback can be provided to the user. A projector could be used to display selected elements on the table. It's also important to show to the user, if the robot is currently executing or processing data, or if it is waiting for user input. A simple three-colored light could be used for this purpose.

For a more expressive user study, additionally the mental/cognitive workload should be measured to define which input modalities are suitable for long-term usage to avoid early exhaustion. [Hirshfield and Solovey, 2009] describes a novel experiment protocol and a set of analysis algorithms that can help UI evaluators or designers of adaptive systems to gain information about the workload experienced by users in the various cognitive resources in their brains, while they work with computer systems.

Cognitive robot cells should also have situational awareness: the choice for a preferred input modality varies for different personalities and situations in which the user finds himself into. Therefore it's important to understand the user itself. If the user is engaged in a task that occupies his hands, he may prefer to use speech input or other input modalities, which don't require moving his hands. Another example is described in [J Coutaz et al., 1995]: suppose that the user wishes to book a flight from somewhere in Europe to Las Vegas. He may not know what the nearest international airport is, so he would prefer to indicate his destination by pointing on a map instead of selecting it from a list. In contrary, if the same user wants to book a flight to a destination where he already knows the international airport code, he would most likely prefer to just write it into a text field instead of selecting the place on a map which is much slower.

# A Questionnaires

This chapter lists all the questions as they were asked during the user study. The participant could choose between English or German questions whereas the questions had still the same meaning.

The user study took about 30 minutes per participant and was divided into the following parts:

1. **Questions:** Background Information (2 Minutes). The user had to fill out personal information like age, gender, expertise in using computers and knowledge about robotics.

2. **Pratical:** Introduction to the system (5 Minutes). The setup of the sensors and touch screen was explained and the user could try out all the available input modalities to get a feeling on how to use them.

3. **Questions:** Expectations (3 Minutes). This part of the questionnaire focused on the expectations of the user regarding the previously presented system. The participant had to order the input modalities according to which input modality he would prefer the most. The possible answers for sorting were randomly mixed to not influence the user by the given order.

4. **Practical:** Hands-On part (15 Minutes). During this step the participant had to teach the robot four different tasks: Pick & Place, Assembly, Point welding, Seam welding. For each task the user had to use all four input modalities (touch, gesture, speech, pen input).

5. **Questions:** Experience & Opinion (5 Minutes). The last part of the questionnaire asked the user about which input modality he preferred (same questions as in the Expectations section, where the user had to order them) and what his opinion on the different input modalities is.

# A.1 English Questionnaire

## A.1.1 Background Information

1. **How old are you?**

   Age: ☐ years old.

2. **What's your gender?**

   ○ male      ○ female

3. **What's your expertise in using computers? I know how to ...**

   ○ Beginner: ... switch it on and open the browser
   ○ Basic: ... use most of the programms installed on my PC and handle files
   ○ Advanced: ... add new hardware and install new software
   ○ Expert: ... program new software and how the computer hardware works

4. **How much do you know about robotics?**

   ○ Not that much: I read/heard about robots but don't know that much about robotics
   ○ Hobby roboticist: I build/program small robots in my freetime
   ○ A lot: I studied robotics, computer science or similar engineering course

5. **Did you already use a TeachPad to program a robot?**

   ○ No, I don't even know what a TeachPad is.
   ○ I know what a TeachPad is, but never programmed a robot with it.
   ○ Yes, I already programmed a robot with a TeachPad

## A.1.2 Expectations

6. **Think about a simple pick and place task, where the robot has to pick up an object from a palette and place it on the table with a specific orientation. What's your estimate in minutes on how long it would take to teach a robot these steps (using current systems used in production)?**

<div style="border:1px solid">       </div> minutes

For each of the following parameters order the input modalities in descending order according to which input modality you would prefer the most. (Most preferred on top, least preferred on bottom)

**7.   Parameter: Select an object**

- ⇕ Touch Input
- ⇕ Pen Input
- ⇕ Point by hand
- ⇕ Speech

**8.   Parameter: Set a location where to place an object**

- ⇕ Touch Input
- ⇕ Pen Input
- ⇕ Point by hand
- ⇕ Speech

**9.   Parameter: Set assembly constraints between two objects**

- ⇕ Pen Input
- ⇕ Speech

**10.  Parameter: Select a point on the object**

- ⇕ Touch Input
- ⇕ Pen Input
- ⇕ Point by hand
- ⇕ Speech

**11.  Parameter: Select an edge on the object**

- ⇕ Touch Input
- ⇕ Pen Input
- ⇕ Point by hand

⬍ Speech

## A.1.3  Experience

**12.  Order the input modalities based on your experienced cognitive load (which modality required the most concentration, where did you have to think the most). Top = high cognitive load, bottom = less cognitive load**

⬍ Touch Input
⬍ Pen Input
⬍ Point by hand
⬍ Speech

Now order the input modalities again.  This time use your experience from previous tasks and order them according to which you preferred the most.  (Most preferred on top, least preferred on bottom)

**13.  Parameter: Select an object**

⬍ Touch Input
⬍ Pen Input
⬍ Point by hand
⬍ Speech

**14.  Parameter: Set a location where to place an object**

⬍ Touch Input
⬍ Pen Input
⬍ Point by hand
⬍ Speech

**15.  Parameter: Set assembly constraints between two objects**

⬍ Pen Input
⬍ Speech

16. **Parameter: Select a point on the object**

   ⬍ Touch Input
   ⬍ Pen Input
   ⬍ Point by hand
   ⬍ Speech

17. **Parameter: Select an edge on the object**

   ⬍ Touch Input
   ⬍ Pen Input
   ⬍ Point by hand
   ⬍ Speech

# A.1.4 Opinion

18. **What are your expectations on the following statements?**

| | a lot easier than expected | | as expected | | much more complicated than I thought |
|---|---|---|---|---|---|
| using the input modalities for programming a robot was | ○ | ○ | ○ | ○ | ○ |
| Touch input was | ○ | ○ | ○ | ○ | ○ |
| Speech input was | ○ | ○ | ○ | ○ | ○ |
| Gesture input was | ○ | ○ | ○ | ○ | ○ |
| Input by pen was | ○ | ○ | ○ | ○ | ○ |
| Overall, using this system to program a robot was | ○ | ○ | ○ | ○ | ○ |

19. **How much do you think is the time saving (in percentage) of using the input modalities compared to using a TeachPad? (Only shown if user knows what a TeachPad, see Question 5)**

   A value of 25% means that using these input modalities requires 75 seconds, when using the TeachPad required 100 seconds.

   [        ] % time saving.

20. **What do you think about the following statements?**

| | strongly disagree | disagree | somewhat disagree | somewhat agree | agree | strongly agree |
|---|---|---|---|---|---|---|
| It's fun to program a robot using multimodality (different types of input) | ○ | ○ | ○ | ○ | ○ | ○ |
| The system should not allow to select a modality. It should pre-select the most suited one for each task. | ○ | ○ | ○ | ○ | ○ | ○ |
| Using the System is natural | ○ | ○ | ○ | ○ | ○ | ○ |
| Speech input is accurate enough for loud industrial environments. | ○ | ○ | ○ | ○ | ○ | ○ |
| I feel more secure if there are less modalities to choose | ○ | ○ | ○ | ○ | ○ | ○ |
| It's intuitive to use speech input | ○ | ○ | ○ | ○ | ○ | ○ |
| It's intuitive to use pen input | ○ | ○ | ○ | ○ | ○ | ○ |
| It's intuitive to use touch input | ○ | ○ | ○ | ○ | ○ | ○ |
| It's intuitive to use gesture input (point by hand) | ○ | ○ | ○ | ○ | ○ | ○ |
| It's intuitive to use keyboard & mouse | ○ | ○ | ○ | ○ | ○ | ○ |
| It was difficult to understand how to use the different modalities | ○ | ○ | ○ | ○ | ○ | ○ |
| Overall, I am satisfied with the ease of completing the tasks in the scenarios | ○ | ○ | ○ | ○ | ○ | ○ |
| Overall, I am satisfied with the amount of time it took to complete the tasks | ○ | ○ | ○ | ○ | ○ | ○ |

21. **Now imagine you are a factory worker and programming industrial robots is your daily job. Answer the following questions:**

| | strongly disagree | disagree | somewhat disagree | somewhat agree | agree | strongly agree |
|---|---|---|---|---|---|---|
| Using the system in my job would enable me to accomplish tasks more quickly | ○ | ○ | ○ | ○ | ○ | ○ |
| Using the system would improve my job performance | ○ | ○ | ○ | ○ | ○ | ○ |
| Using the system in my job would increase my productivity | ○ | ○ | ○ | ○ | ○ | ○ |
| Using the system would slow me down and decrease the efficiency of my job | ○ | ○ | ○ | ○ | ○ | ○ |
| Using the system would make it easier to do my job | ○ | ○ | ○ | ○ | ○ | ○ |
| I would find the system useful in my job | ○ | ○ | ○ | ○ | ○ | ○ |
| Learning to operate the system would be easy for me | ○ | ○ | ○ | ○ | ○ | ○ |
| My interaction with the system would be clear and understandable | ○ | ○ | ○ | ○ | ○ | ○ |
| Using the system would make my job more complicated | ○ | ○ | ○ | ○ | ○ | ○ |
| It would be easy for me to become skillful at using the system | ○ | ○ | ○ | ○ | ○ | ○ |
| I would find the system difficult to use | ○ | ○ | ○ | ○ | ○ | ○ |

## A.2  German Questionnaire

### A.2.1  Hintergrund Informationen

1.  **Wie alt bist du?**

   Alter: [          ] Jahre alt.

2.  **Was ist dein Geschlecht?**

○ männlich      ○ weiblich

**3.   Wie gut kennst du dich mit Computern aus? Ich wei wie man**

○ Anfänger: ... den PC einschaltet und den Browser öffnet
○ Grundkenntnisse: ... Programme auf meinem PC bedient und mit Dateien umgeht
○ Fortgeschritten: ... neue Hardware hinzufügt und neue Software installiert
○ Experte: ... programme entwickelt und wie die Computer-Hardware funktioniert

**4.   Wie viel weißt du über Robotik?**

○ Nicht so viel: Ich habe über Roboter gelesen/gehört kenn mich aber nicht so gut aus
○ Hobby Robotiker: Ich entwickle/programmiere kleine Roboter in meiner Freizeit
○ Ich kenn mich gut aus: ich studiere Robotik, Informatik oder ähnliche Ingenieurswissenschaf

**5.   Hast du schon mal ein TeachPad zur Programmierung eines Roboters verwendet?**

○ Nein, ich weiß nicht mal, was ein TeachPad ist.
○ Ich weiß was ein TeachPad ist, aber habe noch nie einen Roboter damit programmiert.
○ Ja, ich habe bereits einen Roboter mit einem TeachPad programmiert.

## A.2.2  Erwartungen

**6.   Stelle dir eine einfache Pick & Place Aufgabe vor, wo der Roboter ein Teil von einer Palette greifen und dieses auf den Tisch ablegen soll.
Wie lange schätzt du die Zeit, die man benötigt, um diese Aufgabe einem Roboter beizubringen (bei bisher verwendeter Systeme in der Industrie)?**

☐              Minuten

Für jeden folgenden Parameter ordne die Eingabemodalitäten in absteigender Reihenfolge danach, welche du am meisten bevorzugst. (Oben: am meisten bevorzugt, unten: wenigesten bevorzugt)

**7.   Parameter: Objekt auswählen**

‡ Touch Eingabe
‡ Stift Eingabe
‡ Zeigen mit der Hand
‡ Sprache

**8.  Parameter: Position zum Ablegen des Objekts auf dem Tisch**

‡ Touch Eingabe
‡ Stift Eingabe
‡ Zeigen mit der Hand
‡ Sprache

**9.  Parameter: Setzen der Positionen für das Zusammensetzen zweier Objekte**

‡ Stift Eingabe
‡ Sprache

**10.  Parameter: Punkt auf dem Objekt auswählen**

‡ Touch Eingabe
‡ Stift Eingabe
‡ Zeigen mit der Hand
‡ Sprache

**11.  Parameter: Kante auf dem Objekt auswählen**

‡ Touch Eingabe
‡ Stift Eingabe
‡ Zeigen mit der Hand
‡ Sprache

## A.2.3  Erfahrung

**12. Ordne die Eingabemodalitäten basierend auf die erfahrene kognitive Belastung (bei welcher Modalität musstest du am meisten Nachdenken, wo war am meisten Konzentration nötig). Oben = hohe kognitive Belastung, unten = geringe kognitive Belastung**

- Touch Eingabe
- Stift Eingabe
- Zeigen mit der Hand
- Sprache

Jetzt ordne die folgenden Modalitäten nochmal. Diesmal verwende die Erfahrung aus dem praktischen Teil und ordne die Modalitäten basieren auf welche du am meisten bevorzugt hast. (Oben: am meisten bevorzugt, unten: am wenigsten bevorzugt)

**13.  Parameter: Wähle ein Objekt**

- Touch Eingabe
- Stift Eingabe
- Zeigen mit der Hand
- Sprache

**14.  Parameter: Position zum Ablegen des Objekts auf dem Tisch**

- Touch Eingabe
- Stift Eingabe
- Zeigen mit der Hand
- Sprache

**15.  Parameter: Setzten der Positionen für das Zusammensetzen zweier Objekte**

- Stift Eingabe
- Sprache

**16.  Parameter: Punkt auf dem Objekt auswählen**

⁑ Touch Eingabe
⁑ Stift Eingabe
⁑ Zeigen mit der Hand
⁑ Sprache

**17. Parameter: Kante auf dem Objekt auswählen**

⁑ Touch Eingabe
⁑ Stift Eingabe
⁑ Zeigen mit der Hand
⁑ Sprache

## A.2.4 Meinung

**18. Was ist deine Meinung bezüglich folgender Aussagen?**

| | viel einfacher als erwartet | | wie erwartet | | deutlich komplizierter als ich dachte |
|---|---|---|---|---|---|
| verwenden der Eingabemodalitäten zur Programmierung des Roboters war | ○ | ○ | ○ | ○ | ○ |
| Touch Eingabe war | ○ | ○ | ○ | ○ | ○ |
| Spracheingabe war | ○ | ○ | ○ | ○ | ○ |
| Gesten Eingabe war | ○ | ○ | ○ | ○ | ○ |
| Stift Eingabe war | ○ | ○ | ○ | ○ | ○ |
| Insgesamt war die Verwendung des Systems zum Programmieren des Roboters | ○ | ○ | ○ | ○ | ○ |

**19. Wie hoch schätzt du die Zeiteinsparung (in Prozent) bei Verwendung der vorherigen Eingabemodalitäten im Vergleich zur Verwendung eines Teach-Pads? (Nur angezeigt, wenn der Benutzer weiß, was ein TeachPad ist, siehe Frage 5)**

Ein Wert von 25% bedeutet, dass die Eingabemodalitäten 75 sekunden benötigen, wenn die Programmierung mit dem TeachPad 100 Sekunden dauerte.

| | % Zeitersparnis.

**20. Was denkst du zu den folgenden Aussagen?**

| | stimme über-haupt nicht zu | stimme nicht zu | stimme eher nicht zu | stimme eher zu | stimme zu | stimme voll-kommen zu |
|---|---|---|---|---|---|---|
| Programmieren des Roboters mit multimodaler Eingabe macht Spaß | ○ | ○ | ○ | ○ | ○ | ○ |
| Das System sollte keine Auswahl an Modalitäten bieten, sondern nur die am Besten geeignete anbieten | ○ | ○ | ○ | ○ | ○ | ○ |
| Die Verwendung des Systems ist natürlich | ○ | ○ | ○ | ○ | ○ | ○ |
| Spracheingabe ist genau genug in lauter industrieller Umgebung | ○ | ○ | ○ | ○ | ○ | ○ |
| Ich fühle mich sicherer, wenn weniger Eingabemodalitäten zur Auswahl stehen | ○ | ○ | ○ | ○ | ○ | ○ |
| Spracheingabe ist intuitiv | ○ | ○ | ○ | ○ | ○ | ○ |
| Stift-Eingabe ist intuitiv | ○ | ○ | ○ | ○ | ○ | ○ |
| Touch-Eingabe ist intuitiv | ○ | ○ | ○ | ○ | ○ | ○ |
| Gesten Eingabe (Zeigen mit der Hand) ist intuitiv | ○ | ○ | ○ | ○ | ○ | ○ |
| Tastatur & Maus ist intuitiv | ○ | ○ | ○ | ○ | ○ | ○ |
| Es war schwer, die verschiedenen Modalitäten zu verstehen | ○ | ○ | ○ | ○ | ○ | ○ |
| Insgesamt bin ich mit der Leichtigkeit der Ausführung der Aufgaben in den Szenarien zufrieden | ○ | ○ | ○ | ○ | ○ | ○ |
| Insgesamt bin ich mit der benötigten Zeit zur durchführung der Aufgaben zufrieden | ○ | ○ | ○ | ○ | ○ | ○ |

21. **Jetzt stell dir vor du arbeitest in einer Fabrik und deine tägliche Aufgabe ist es, Industrieroboter zu programmieren. Beantworte folgende Fragen:**

| | stimme über­haupt nicht zu | stimme nicht zu | stimme eher nicht zu | stimme eher zu | stimme zu | stimme voll­kommen zu |
|---|---|---|---|---|---|---|
| Verwenden des Systems in meiner Arbeit würde mir er­möglichen, Aufgaben schnel­ler zu erledigen | ○ | ○ | ○ | ○ | ○ | ○ |
| Die Verwendung des Systems würde meine Arbeitsleistung erhöhen | ○ | ○ | ○ | ○ | ○ | ○ |
| Die Verwendung des Systems würde meine Produktivität steigern | ○ | ○ | ○ | ○ | ○ | ○ |
| Die Verwendung des Systems würde meine Arbeit verlang­samen und die Effizienz ver­mindern | ○ | ○ | ○ | ○ | ○ | ○ |
| Die Verwendung des Systems würde meinen Job einfacher machen | ○ | ○ | ○ | ○ | ○ | ○ |
| Ich würde das System nütz­lich finden | ○ | ○ | ○ | ○ | ○ | ○ |
| Lernen das System zu ver­wenden wäre einfach für mich | ○ | ○ | ○ | ○ | ○ | ○ |
| Die Interaktion mit dem Sys­tem wäre klar und verständ­lich | ○ | ○ | ○ | ○ | ○ | ○ |
| Das System würde meine Ar­beit komplizierter machen | ○ | ○ | ○ | ○ | ○ | ○ |
| Es wäre einfach für mich, ein Experte in dem System zu werden | ○ | ○ | ○ | ○ | ○ | ○ |
| Ich fände das System kompli­ziert zu bedienen | ○ | ○ | ○ | ○ | ○ | ○ |